\documentclass[journal]{IEEEtran}

\usepackage{cite}

\usepackage{graphicx}          

\usepackage{mathtools}

\DeclarePairedDelimiter{\ceil}{\lceil}{\rceil}

\usepackage{listings}

\lstdefinelanguage{customc}{
  morekeywords={INIT},
  sensitive=false,
  morecomment=[l]{//},
  morestring=[b]'',
}

\lstdefinestyle{kpnapi}{
  language=customc,
  basicstyle=\small\ttfamily,
  keywordstyle=\color{black}\bfseries,
  escapechar=@,
  tabsize=2,
  breaklines=true,
  frame=trbl,
  captionpos=b
}

\usepackage{comment}
\usepackage[dvipsnames]{xcolor}
\usepackage{footmisc}

\usepackage{flushend}
\usepackage[frozencache]{minted}
\usepackage{amssymb}
\usepackage{pifont}

\newcommand{\marktick}{\textcolor{OliveGreen}{\ding{51}}}%
\newcommand{\markcross}{\textcolor{red}{\ding{55}}}%

\newcommand{\asrev}[1]{{#1}}
\newcommand{\fcrev}[1]{{#1}}

\newcommand{\loopname}[1]{\texttt{\textbf{#1}}}

\newenvironment{definition}[1][Definition]{\begin{trivlist}
\item[\hskip \labelsep {\bfseries #1}]}{\end{trivlist}}

\begin{document}

\title{Optimally Scheduling CNN Convolutions for Efficient Memory Access
}


\author{Arthur~Stoutchinin,
Francesco~Conti,~\IEEEmembership{Member,~IEEE,}
Luca~Benini~\IEEEmembership{Fellow,~IEEE,}
%
\thanks{A. Stoutchinin is with STMicroelectronics France, 12 Rue Jules Horowitz, 38019 Grenoble, France (email: arthur.stoutchinin@gmail.com).}
\thanks{F. Conti and L. Benini are with the Integrated System Laboratory of ETH Z\"urich, ETZ, Gloriastrasse 35, 8092 Z\"urich, Switzerland (e-mail: dpalossi@iis.ee.ethz.ch, fconti@iis.ee.ethz.ch, eflamand@iis.ee.ethz.ch, lbenini@iis.ee.ethz.ch); and also with the Department of Electrical, Electronic and Information Engineering of University of Bologna, Viale del Risorgimento 2, 40126 Bologna, Italy (e-mail: f.conti@unibo.it, luca.benini@unibo.it).}
\thanks{This work has been funded in part by project EC Horizon-2020 ALOHA~(g.a.~780788).}%
}


\markboth{IEEE TRANSACTIONS ON COMPUTER-AIDED DESIGN OF INTEGRATED CIRCUITS AND SYSTEMS}%
{??}
%

\maketitle

\begin{abstract}

Embedded inference engines for convolutional networks must be parsimonious
in memory bandwidth and buffer sizing to meet power and cost constraints.
We present an analytical memory bandwidth model for
loop-nest optimization targeting architectures with application managed
buffers. We applied this model to optimize the CNN convolution loop-nest.
We show that our model is more accurate than previously published models.
Using this model we \fcrev{can} identify a non-trivial dataflow schedules that result in
lowest communication bandwidth given a tight local
buffering constraints. We show that optimal dataflow schedules are
implementable in practice and that our model is accurate with respect to
a real implementation; \fcrev{moreover, we introduce an accelerator architecture, named Hardware Convolution Block (HWC), which implements the optimal schedules, and we show it achieves up to 14$\times$ memory bandwidth reduction compared to a previously published accelerator with a similar memory interface, but implementing a non-optimal schedule.} 

\end{abstract}


%
\IEEEpeerreviewmaketitle

\section{Introduction}

Convolutional neural networks (CNNs) are widely
used for solving artificial intelligence problems, such as object
and voice recognition, scene  labeling and others~\cite{LeCunBenHin2015}.
Many research and development efforts have recently focused on
domain-specific CNN
accelerators in order to meet the high computational requirements of
CNN inference with reasonable energy and cost efficiency~\cite{SzeCYE17}.

\fcrev{One of the primary bottlenecks} for implementing efficient and cost effective embedded
CNN accelerators
for state-of-the-art deep convolutional networks is the memory system.
Large volume of data accessed (weights) and produced (feature maps)
during CNN inference computation makes it
difficult to simultaneously buffer the input
feature maps, the output feature maps, and the filter weights
in limited 
internal accelerator memory.
One way to alleviate this issue is to use large
SRAM buffers (up to a few MBytes may be used)
in order to completely eliminate main memory traffic ~\cite{ChenKES17,DuAl2015}.
When massive 
accelerator area budget is available,
this may be an acceptable approach. However, large amounts of memory
are not affordable in deeply-embedded markets, such as mobile or
IoT clients, for example.

While completely absorbing all CNN memory traffic in internal accelerator
storage is usually not possible, the memory bandwidth requirement for a given
accelerator storage capacity can be significantly reduced if sufficient data
reuse happens.
The data reuse pattern, in time and space, is determined by
the {\it dataflow schedule} of
computation.
Generally, the efficiency and performance impact of the dataflow schedule
varies with CNN topology and size
making it difficult to adapt the accelerator architecture to different
CNNs.



Existing work on scheduling CNN
computations~\cite{PeemenMC2013,ZhangLSGXC15,YangPRBRKRPH16} is based on
memory models originally developed for cache-based memory
hierarchies~\cite{WolfLam91}.
Previously published models essentially search the tiling (or blocking)
space of the CNN
convolution loop-nest with the objective to identify the innermost loops
set such that the working set of these innermost loops fits the available
internal storage while the data transfers between the internal and
external memories are minimized.
However, existing models have not been adapted to \fcrev{explicitly} application managed
buffers, which \fcrev{constitute by far} the most
common memory architecture template for CNN
accelerators~\cite{PeemenAl2013,GokhaleJDMC14,ContiBenini15,ChenCXST16,SimPKBCK16,ChenKES17,Nervana,GoogleTPU,ImagPowerVR,DesoliAl16}.
In this case, published models overestimate internal storage requirements
of the CNN computation and result in sub-optimal dataflow schedules.


In this paper, \fcrev{we provide three key contributions to the state-of-the-art. First, we} propose a new analytical memory performance model to evaluate dataflow schedules in terms of their \fcrev{local memory requirements and overall external (off-accelerator) memory traffic.}
\fcrev{We compared the best dataflow schedules that can be identified with our model with the current state-of-the-art models~\cite{PeemenMC2013}, showing that they require 5-15\% less memory traffic when applied to a number of state-of-the-art CNN topologies, enabling a better usage of available resources.}
\fcrev{Moreover, to validate our model, we applied it on the case study of the design of}  a flexible CNN
accelerator for deeply embedded Systems-on-Chip.
Our accelerator architecture is based on a shared memory cluster, similar
to \cite{Rossi0MPLGTCFB15}, enhanced with dedicated convolution hardware
units (HWC). 

We have used our analytical memory bandwidth model
to perform architectural exploration of the HWC accelerator
with the objective to find a dataflow schedule that results in the best
trade-off between the capacity of the
hardware unit storage and the \fcrev{traffic to} shared memory.
The dataflow schedule found using our methodology is non-trivial and it
achieves \fcrev{up to 10-14$\times$ reduction} with respect to a previously published
accelerator based on a similar architecture~\cite{ContiBenini15}, \fcrev{while assuming similar amounts of internal (in-accelerator) memory}.
\fcrev{In absolute terms, a complete HWC implemented from the dataflow schedule by means of high-level synthesis can sustain a throughput of up to} 16 Multiply-Accumulate (MAC) operations
per clock cycle \fcrev{using only 1KB of internal storage and having only 3$\times$32-bit ports to off-accelerator shared memory.}
%
\asrev{Finally},
%
%
\fcrev{we verify the accuracy of our memory model} by comparing \fcrev{the predicted memory traffic} with the real number of memory accesses measured
from our accelerator implementation\fcrev{, showing a maximum deviation of 0.5\% in a few corner cases.}

The rest of the paper is organized as follows:
\asrev{
Section \ref{background} introduces the CNN convolution notations, the
opportunities and challenges of application-managed buffers with respect
to caches, and the CNN data locality optimization problem; Section \ref{related}
explains previously published related work;
}
Section
\ref{model} presents our analytical memory performance model applied to
the CNN convolution computation and compares
it to previously published models; \asrev{finally,} Section \ref{architecture} illustrates
application of this model for the development of a shared memory
CNN convolution hardware accelerator.


\section{Background and notation}
\label{background}

\begin{figure}[t]
\centering
\includegraphics[width=0.85\columnwidth]{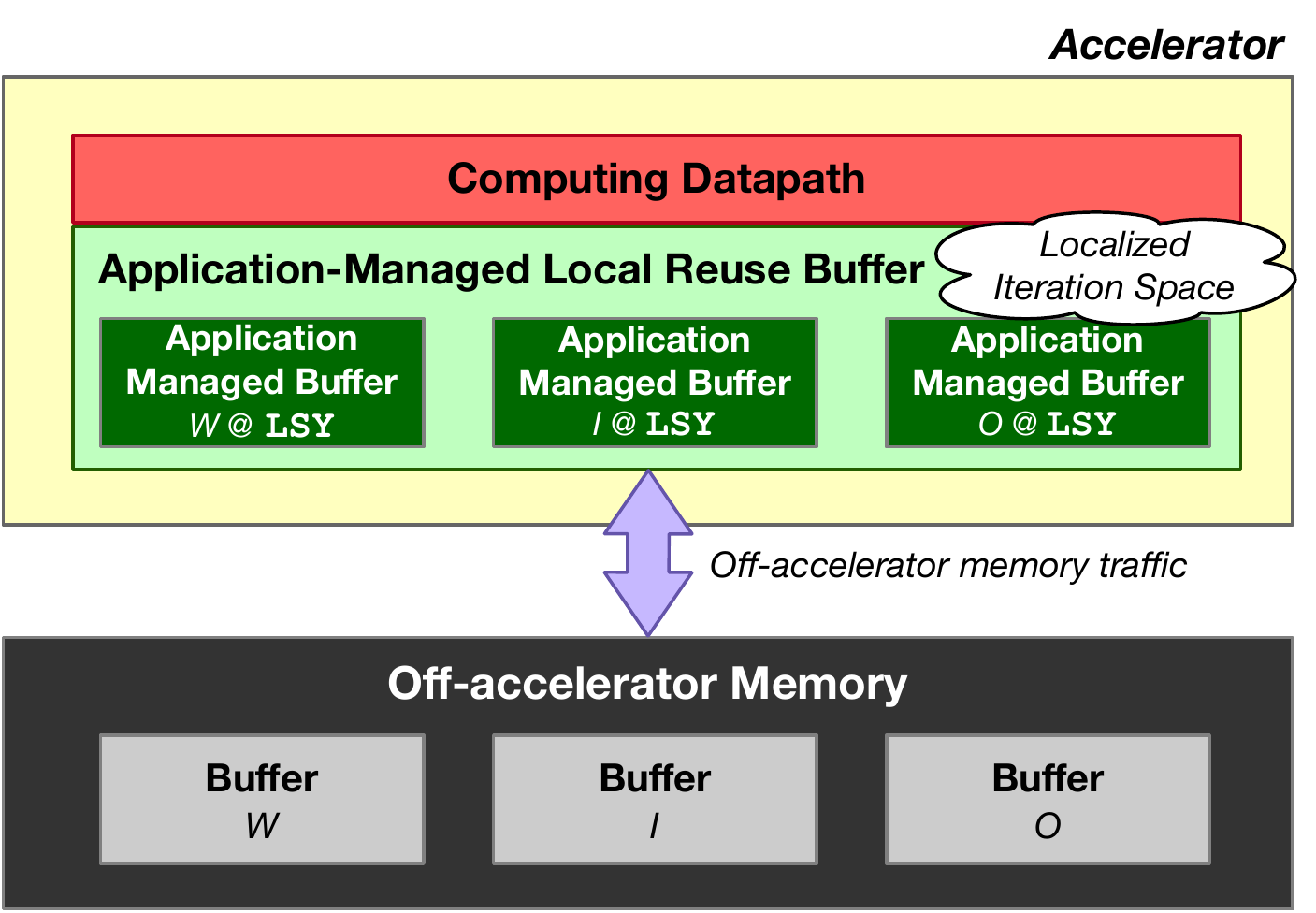}
\caption{\label{Fig14} \fcrev{Generic view of a CNN accelerator combining a computing datapath with an optimized application-managed \textit{local reuse buffer}, and off-accelerator memory external to the datapath.}}
\end{figure}

\lstset{style=kpnapi,numbers=left,stepnumber=1,escapechar=@}
\usemintedstyle{friendly}

\begin{figure}[t]
\centering
\begin{minted}[numbersep=5pt, gobble=0, frame=lines, fontsize=\footnotesize, framesep=2mm, mathescape=true, escapeinside=||]{C}
  // M output fmaps loop
  LOF: for (m = 0; m < M; m++)
    // C input fmaps loop
    LIF: for (c = 0; c < C; c++)
      // spatial loops (ExE)
      LSY: for (y = 0; y < E; y++)
        LSX: for (x = 0; x < E; x++)
          // filter loops (RxR, stride S)
          LFY: for (k = 0; k < R; k++)
            LFX: for (l = 0; l < R; l++)
            {
              p = I[c][y*S+k][x*S+l];
              w = W[m][c][k][l];
              O[m][y][x] += p*w;
            }
\end{minted}
\caption{\label{Fig2} Canonical form of the CNN convolution layer loop-nest.}
\end{figure}

In this paper we propose an analytical model for evaluating
storage requirements and memory bandwidth of a generic CNN convolutional layer,
expressed as a 6 level loop nest, as shown in Figure \ref{Fig2} \fcrev{in a ``canonical'' form}.
Table \ref{Tab2} lists symbols used in discussing the CNN convolutional layer
and their meaning.
In the Figure \ref{Fig2}, there are 3 arrays referenced in the loop-nest:
 $I$ holds the
$C$ input feature maps of size $H \times H$;
$W$ holds $M \times C$ convolution
kernels, with $R \times R$ weights each; and $O[M][E][E]$ holds $M$ output
feature maps of size $E \times E$\footnote{We use square feature maps to
  avoid overburdening
  the mathematical notation, although the model can be easily extended to
  general rectangular feature maps.}.

\begin{table}[t]
\centering
\begin{tabular}{|c|l|}
\hline
\textbf{Symbol}   &  \textbf{Description}  \\
\hline
$H$   & Input feature map size ($H\times H$)   \\
\hline
$E$   & Output feature map size ($E\times E$)  \\
\hline
$R$   & Convolution kernel size ($R\times R$)  \\
\hline
$S$   & Convolution kernel stride \\
\hline
$C$   & Number of input feature maps  \\
\hline
$M$   & Number of output feature maps \\
\hline
\end{tabular}
\vspace{5pt}
\caption{\label{Tab2} Symbols used in this paper}
\end{table}

We consider the problem of optimizing the CNN convolutional layer
memory access pattern for a two-level memory hierarchy.
Figure \ref{Fig14} shows a generic view for
an accelerator consisting of computing datapath, 
{\it local reuse buffer} optimized for this datapath,
and 
{\it off-accelerator memory} external to the computing datapath.
The problem consists in minimizing the number of memory accesses to the
off-accelerator memory given
a limited local buffer capacity.

\subsection{Data reuse in CNN convolutional layers}

Data reuse occurs when a reference within a loop accesses the same
data element in different iterations.
Convolutional loop-nest shown in Figure \ref{Fig2} contains several
opportunities for data reuse:
\begin{itemize}
\item {\bf convolution reuse}: Each input feature map pixel is reused
  $R^2$ times within each input feature map.
\item {\bf weight reuse}: Each kernel weight is reused $E^2$
  times.
\item {\bf input fmap reuse}: Each input feature map pixel is reused across
  $M$ output feature map computations.
\item {\bf output fmap accumulation}: Each output feature map pixel is reused
  across accumulations of partial results from $C$ input feature maps.
\end{itemize}

\begin{table}[tb]
\centering
\begin{tabular}{|r|c|c|c|c|c|c|}
\hline
 \textbf{Carries?} & \loopname{LFX} & \loopname{LFY} & \loopname{LSX} & \loopname{LSY} & \loopname{LIF} & \loopname{LOF} \\
\hline
$I$    &   \marktick{} or \markcross  & \markcross{} or \marktick &  \marktick{} or \markcross  & \markcross{} or \marktick & \markcross & \marktick \\
$W$    &   \markcross  &  \markcross  &  \marktick  & \marktick & \markcross & \markcross \\
$O$    &   \marktick  &  \marktick & \markcross & \markcross & \marktick & \markcross \\
\hline
\end{tabular}
\vspace{5pt}
\caption{\label{TabCarry} Carrying loops of array references of the CNN loop-nest.}
\end{table}

We say that the reuse of a reference is {\it carried} by a loop if the
same memory
location is used by different iterations of that loop~\cite{LamRotWolf91}.
In Figure \ref{Fig2}, reuse of the $W$ 
reference is {\it carried} by two loops, \loopname{LSX} and \loopname{LSY}; reuse of the
$O$ reference is carried by the loops  \loopname{LFX},\loopname{LFY}, and  \loopname{LIF}. 
The reuse of the $I$ reference is slightly more complex because it is
carried by a combination of loops:
which pair of loops between (\loopname{LFX},\loopname{LSX}) or (\loopname{LFY},\loopname{LSY}) is
carrying the reuse of array $I$ depends on relative ordering of the loops; the
reuse is carried by the outer loop in each pair.
The $I$ reference is also carried by loop \loopname{LOF}.
\fcrev{
Table~\ref{TabCarry} resumes which loops are carrying each array reference in the CNN loop-nest.
}
%

In order to take full advantage of the reuse, such that every
data is \fcrev{loaded from/stored to off-accelerator} memory only once,
large local buffering capacity is necessary.
In Figure \ref{Fig2}, the entire set of input feature
maps needs to be stored in the local buffer to reuse the
input data across the loop \loopname{LOF}. To reuse the
accumulated partial output feature maps $O$ across loop \loopname{LIF},
one full output feature map needs to be stored in local buffer.
To put this into perspective, the total amount of buffering required
for the second
convolution layer of AlexNet~\cite{Krizhevsky2012}, with
$H=55, E=27, C=96, M=256, R=5$, \asrev{exceeds} \fcrev{284~kB} if we consider
that each element in the input and output feature maps and the kernel
weights is 1 byte in size.

If \fcrev{no large local buffer is available}, optimal data reuse cannot be
achieved and some data must be accessed from the external level in memory
hierarchy multiple times.
\asrev{From Figure \ref{Fig2}},
if only $R$ lines of input feature maps can be
buffered locally, similar to 2D convolvers line buffers~\cite{FarabetPHL09},
then every input feature map pixel will
need to be re-loaded $M$ times, once for each iteration of loop \loopname{LOF}.
\asrev{For the second convolution layer of the AlexNet, this would
multiply the required number of memory accesses for the
array $I$ by 256}.
The problem is that, although there is plenty of data reuse in the CNN
loop-nest, unless the entire working set fits the local buffer, this
reuse cannot be taken full advantage of.
The remainder of this Section builds up the necessary background and notation
to tackle the problem of maximizing data reuse \fcrev{in local memory} in an analytical fashion.

\subsection{Local Reuse Buffers}

\fcrev{A generic 2-level memory hierarchy such as that exemplified in Figure~\ref{Fig14} exposes two memory levels: \textit{off-accelerator memory}, where we assume all data involved in the CNN loop-nest is resident, and a \textit{local reuse buffer} that usually cannot host the entirety of the data, but is vastly faster and more energy-efficient than \fcrev{off-accelerator} memory.
We assume the local reuse buffer to be implemented either as a data cache or as a software-managed scratchpad memory.
This buffer is used to host data that is reused multiple times.
%
%
While data reuse is inherent in the computation and not
dependent on a particular shape of the loop-nest, the usage of a 
local reuse buffer of any kind implies that the reuse only translates
into a reduction of memory accesses if there
is enough \textit{data locality}, i.e. if data inside the local buffer are
reused within a short period of time and are not replaced between reuse
accesses.
}

\subsubsection{Data caches}
Existing reuse evaluation methods derive from cache behavior and
\asrev{
build a \textit{localized iteration space}, i.e. a set of $k$ innermost loops
of a loop-nest where data locality is exposed~\cite{WolfLam91}.
It is assumed
}
that all array references inside the {\it localized iteration space}
\asrev{need to be} simultaneously stored in \fcrev{a} local
reuse buffer~\cite{LamRotWolf91,PeemenAl2013,ZhangLSGXC15,YangPRBRKRPH16}.
Indeed,
in the context of a data cache, every array reference is mapped to a unique
location in the cache.
If the cache capacity is smaller then
required for holding all array references,
reused data can be displaced from the cache and
are not guaranteed to remain in the cache in every iteration that they are
referenced.
Thus, all data touched inside the \textit{localized iteration space} need to
be cached in order
to benefit from the data reuse.
\fcrev{
For example, in order to reuse elements of array $I$ across the loop \loopname{LSY} in
Figure \ref{Fig2}, the \textit{localized iteration space}
will have to include loops \loopname{LFX}, \loopname{LFY}, \loopname{LSX} and \loopname{LSY}.
The data cache would need to hold $H \times H$ elements of $I$ - otherwise
referencing $I$ could displace some other reused data, $O$ for example.
The data cache would also need to hold $E \times E$ elements of
the output feature map array $O$ - otherwise, referencing the array $O$ may
displace elements of array $I$ from the cache before they have been reused.
}

\subsubsection{Application-managed scratchpad memories}

Dedicated hardware accelerators, including CNN accelerators, commonly use
application-managed scratchpad memories \fcrev{as local reuse buffers} instead of caches,
\fcrev{as these are deemed to provide better} performance,
predictability, and energy efficiency~\cite{ContiAl2014,SzeCYE17}.
\fcrev{In this case, data placement, reuse, and transfer have to be managed
explicitly by partitioning the local reuse buffer in a set of \textit{application-managed
buffers}, one per each array referenced in the loop-nest.
Application-managed buffers can be partitioned statically (i.e. by using physically
separate memory to implement them) or dynamically (i.e. by partitioning a single piece of
memory so that all array references fit).
}
%
\asrev{Instead of a single \textit{localized iteration space}, each array
reference can have its own data locality scope. Thus,
}
\fcrev{
utilization of the local reuse buffer can be optimized
by choosing, for each array reference, a nested loop level at which data
are buffered for reuse.
We call this level the \textit{buffering level} of the array reference
}.

\asrev{
The number of loop iterations, $d$, that separates two consecutive accesses
to the same data element, is called the \textit{dependence distance of data
 reuse} or simply the \textit{reuse distance}~\cite{WolfLam91}.
Only the elements of the array touched by $d$ loop iterations need to be
buffered in application-managed local buffer in order to ensure that reused
data is preserved across loop iterations.
}
%
%
\asrev{
For example, in Figure \ref{Fig2}, the reuse distance of the reference to
array $I$ across the loop \loopname{LSY} is $H \times R$ iterations
($H$ elements of $I$ are touched iterating over the \loopname{LSX} loop).
Therefore, if an implementation chooses to buffer the array $I$ at the
\loopname{LSY} level, enabling $I$ data reuse across this loop, $H \times R$
elements of $I$ need to be buffered in the local buffer.
Similarly, buffering the array $O$ at the \loopname{LFY} level, i.e.
reusing array $O$ data across the two innermost loops, requires only a
single element of the array $O$ to be buffered inside the local buffer.
}

\fcrev{In the following of this work, we will focus mostly on the case of application-managed scratchpad memories, which are used in the majority of computing platforms dedicated to CNNs ~\cite{PeemenAl2013,GokhaleJDMC14,ContiBenini15,ChenCXST16,SimPKBCK16,ChenKES17,Nervana,GoogleTPU,ImagPowerVR,DesoliAl16}.}

\subsection{Data Locality Optimization}

\lstset{style=kpnapi,numbers=right,stepnumber=1,escapechar=@}

\begin{figure}[bt]
\centering
\begin{minted}[numbersep=5pt, gobble=2, frame=lines, fontsize=\footnotesize, framesep=2mm]{C}
  // output fmaps -- loop on tiles
  LTOF: for (mm = 0; mm < M; mm += mss)
    // input fmaps -- loop on tiles
    LTIF: for (cc = 0; cc < C; cc += css)
      // spatial -- loops on tiles
      LTSY: for (yy = 0; yy < E; yy += iss)
        LTSX: for (xx = 0; xx < E; xx += jss)
          // output fmaps -- tile loop
          LOF: for (m=mm; m<min(mm+mss,M); m++)
            // input fmaps -- tile loop
            LIF: for (c=cc; c<min(cc+css,C); c++)
              // spatial -- tile loops
              LSY: for (y=yy; y<min(yy+iss,E); y++)
                LSX: for (x=xx; x<min(xx+jss,E); x++)
                  // kernel -- tile loops
                  LFY: for (k=0; k<R; k++)
                    LFX: for (l=0; l<R; l++)
                    {
                      p = I[c][y*S+k][x*S+l];
                      w = W[m][c][k][l];
                      O[m][y][x] += p*w;
                    }
\end{minted}
\caption{\label{Fig3} Tiled CNN convolution layer loop-nest.}
\end{figure}

\asrev{
Memory performance optimization methods, such as \cite{LamRotWolf91}
or \cite{PeemenAl2013}, use \textit{reordering} and
\textit{tiling} of the loop-nest to maximize the data locality.
Loop reordering places a subset of the $k$ loops that form the
\textit{localized iteration space} at the innermost position.
Loop tiling partitions the loop-nest iteration space into a number of
smaller blocks, such that data used inside the
\textit{localized iteration space} stays in the local buffer
until it is reused.
Figure \ref{Fig3} shows a general form of tiled convolution loop-nest,
where 
the choice of
tile sizes $mss, css, iss, jss$ leads to different data locality results.
}

\asrev{
With caches, the order in which loops from the
{\it localized iteration space} execute is not important because 
the ability to reuse data depends solely on the total number of data elements loaded between reuses, i.e. only on the size of the
{\it localized iteration space}.
Conversely, with application-managed buffers, the order of loop execution has a
substantial effect on data locality.
To see this, consider reusing elements of array $I$ across the \loopname{LIF}
loop in the Figure \ref{Fig3}. 
Let $h$ be the number of elements of $I$ touched across iterations of each
of \loopname{LSX} and \loopname{LSY}. Since the reuse distance of $I$ with
respect to the loop \loopname{LSY} is $R$ iterations, and there is no reuse
across the loop \loopname{LIF},
$h \times R$ elements need to be buffered in the local
buffer to ensure the reuse of $I$ across the loop \loopname{LIF}.
Consider reordering loops \loopname{LIF} and \loopname{LOF}.
The reuse distance of $I$ with respect to the loop \loopname{LOF} is $mss$,
therefore the number of elements touched by $mss$ iterations of this loop
need to be buffered in order to ensure the reuse of $I$ across the loop
\loopname{LIF}. A local buffer of $h \times h$ elements of $I$ is thus
necessary.
}

\asrev{
Generally, with limited buffering capacity it is not possible to fully exploit the full amount of data reuse for all array references at the same time.
Different choices of loop order, tile sizes, and buffering levels lead to dataflow schedules with different reuse characteristics.
%
\fcrev{
In the following Sections, we first review the related work in the state-of-the-art, with previously proposed models to evaluate and compare dataflow schedules; then we introduce our proposed analytical memory performance model for evaluating this tradeoff in application to the CNN convolution loop-nests.
}}

\section{Related Work}
\label{related}

The main difficulty in implementing cost effective embedded
CNN accelerators is optimizing memory organization to
minimize off-chip memory transfers and perform them as efficiently
as possible.

\asrev{
One research direction that aims at alleviating
the CNN memory bottleneck is data compression. Various data compression
techniques have been proposed in literature:
quantization~\cite{GuptaAGN15,Gysel16,WuLWHC16} and 
binarization~\cite{CourbariauxBD15,CourbariauxB16,RastegariORF16,AndriCRB18,ContiSchBen2018},
data compression~\cite{HanLMPPHD16}.
A survey on CNN data compression can be found in \cite{CheWan17}.
Our work is orthogonal to these techniques and can be used on top of them.
Indeed, in our implementation we have used a dynamic fixed-point data
quantization technique from \cite{Gysel16}.
We shall also point out that recently sparse neural networks emerged as
one solution to reduce the amount of computation and memory required for
the CNN processing~\cite{liu2015sparse,ChangpinyoSZ17,Zhouetal2018,CavigelliExtendedBitPlaneCompression2018}.
This approach is beyond the scope of our work.
}

The straightforward approach
\asrev{
to scheduling CNN computations
}
is to use
the 2D convolution along with line buffers for the
data reuse~\cite{FarabetPHL09,FarabetAl10,MerollaAAIMM11,GokhaleJDMC14,ContiBenini15,QadeerHSVKH15,AzarkhishRLB17}.
Although simple to implement, \asrev{such} accelerators are often limited to a
particular size of convolution kernels and input image size.
Additionally, they can only exploit a
fraction of data reuse that exists inside the CNN convolution computations.

Several architectures attempted to overcome the 2D convolution limitations
by scheduling the CNN convolution dataflow
\asrev{
differently, 
with the \textit{dataflow schedules} being a result of ad-hoc
and empirical exploration.
}
For example, the Eyeriss accelerator~\cite{ChenES16,ChenKES17} proposed the
{\it row stationary dataflow}
which schedules the CNN convolution
on a 2D array of processing elements and optimizes the data reuse
by exploiting the low-cost memory levels, the PE scratchpads and
the inter-PE communication, while minimizing data accesses to the high-cost
levels, including the large on-chip global buffer and the off-chip DRAM.
Another CNN accelerator, Angel-Eye~\cite{GuoSQYWYHWY18}, is a FPGA
accelerator that combines multiple
parallel 2D convolvers
such that several partial sums are accumulated 
simultaneously.
%
%
The DianNao architecture~\cite{ChenDSWWCT14}
have determined 
dataflow schedule and buffer sizes
experimentally.
The authors acknowledged that, like many processing architectures, DianNao's
efficiency and scalability remained severely limited by memory bandwidth
constraints.
DianNao's successor, ShiDianNao accelerator~\cite{DuAl2015} was directly
integrated with an image processor sensor, therefore fully eliminating
main memory.
This approach is not scalable as only a few small CNNs can be accommodated.
Another successor, DaDianNao~\cite{ChenLLZHWLCXST14} employed a sufficiently
large on-chip eDRAM for storing large CNNs close to the datapath.
The DLAU architecture~\cite{WangGYLXZ17} utilizes tiling techniques and
minimizes 
main memory transfers by accumulating a few
partial results in 
internal buffers.

%
\fcrev{To solve the same problem in a more formal way, several} publications 
proposed analytical memory performance models.
\asrev{For example,}
memory optimization models for stencil computational kernels were
published in \cite{RanaBBNAS16} and in \cite{CongLXZ16}. Stencils differ
from CNN convolutional layers in that they do not need
to handle a large amount of convolution kernel weights. Therefore,
these models cannot be used to optimize the CNN computation.
TETRIS~\cite{GaoPYHK17} analytically derived optimal {\it dataflow schedules} 
for
the CNN convolution by simplifying the problem. They proposed the
{\it bypass ordering} where internal on-chip storage is bypassed for
two out of the three input streams in the
convolution layer using a large register file for buffering the {\it bypassed}
two streams instead. Bypass ordering is significantly
simpler than the general computation scheduling problem,
and it is possible to analytically derive the optimal loop-nest shape
without recurring to exhaustive search of the solution space.
However, the bypass ordering relies on
a particular architecture and requires large local register
file and buffer.


Wolf and Lam~\cite{WolfLam91,LamRotWolf91} used 
loop blocking and reordering techniques 
to capture data locality and reduce the memory traffic in
scientific computations.
They used a combination of loop
interchange, skewing, and reversal
(unimodular transformations) with loop tiling (blocking) to
improve data locality of loop-nests
in the context of
memory hierarchies with caches.
Such model is inaccurate in the context of memory hierarchies with
application-managed buffers. 
%
%
%
As a result, Wolf's method overestimates the real buffering requirements
and required memory bandwidth
of a computation and leads to sub-optimal {\it dataflow schedules}.

Peemen {\it et al.}~\cite{PeemenMC2013,PeemenAl2013} proposed an architecture
model
where a computation loop nest is split into two parts: an 
\asrev{\textit{innermost tile}, (similar} to M.Wolf's
{\it localized iteration space}),
for execution on the accelerator, and outer
{\it controlling loops} that run on a host processor.
\asrev{
Peemen's approach improves on M.Wolf's cache model significantly by taking into account that with application
managed buffers some data can be reused between consecutive executions of the
\textit{innermost tiles}.
}
They proposed a design flow for selecting the best {\it dataflow
schedule} to maximize data reuse given a buffer size restriction. This is
achieved by tiling of the CNN convolution loop-nest \asrev{and reordering 
the controlling loops}.
\asrev{
However, as explained in section \ref{model}, the buffer estimation remains inaccurate and the
solution is sub-optimal.
}

Zhang {\it et al.}~\cite{ZhangLSGXC15} proposed an analytical approach for
analyzing computing throughput and required memory bandwidth of a CNN design
on an FPGA platform.
\asrev{
Similar to Peemen's method, the CNN loop-nest is tiled with the
innermost tiled loops executing in the FPGA, while controlling loops
execute in the host.
Data need to be loaded into (and read from) the FPGA internal memory for
each execution of the \textit{innermost tile}.
In order to reduce the main memory traffic they use local memory
promotion~\cite{PouchetZSC13} for placing out-of-FPGA communication operations
optimally.
The local memory promotion allows moving a transfer of
data from array $X$ across the innermost controlling loop, when this loop
carries full reuse of $X$, i.e. this loop's iterator does not appear in any
reference to array $X$.
Similarly, to Peemen's method, Zhang \textit{et al.} explore 4 possibilities
for the 4 different innermost controlling loops in the convolution loop-nest.
The resulting \textit{dataflow schedules} are less efficient than in
Peemen's approach because only a subset of array references benefit from
data reuse across consecutive executions of the innermost tile.
}

\asrev{
Yang {\it et al.}~\cite{YangPRBRKRPH16} published a method for the
convolution loop-nest tiling for multi-level memory hierarchies.
The authors focus on improving the total energy consumption in such
systems.
In Yang's work,
one loop \textit{blocking} is performed for each target memory hierarchy
level, building one
{\it localized iteration space} per memory level.
Blocking the convolution loop-nest at each level can be though of as tiling
the loops in the loop-nest, and then exchanging the order in which the
controlling loops are executed.
Yang \textit{et al.} acknowledge that optimal application of their algorithm
to multiple memory hierarchy levels is quite computationally costly. 
In order to achieve a reasonable computation time, instead of optimal
multi-level blocking, the authors then propose to apply a 2-level blocking
repeatedly starting from lower memory hierarchy level to the upper, while
adjusting the lower level results at each new level.
However, for the 2-level blocking, since the memory level buffering requirements are estimated as the
sum of all data elements touched inside the level's
\textit{localized iteration space}, this approach results in
\textit{dataflow schedule} quality
essentially similar to the Peemen's method.
}

Overall, derived from the cache behavior, the above memory performance models assume
that \asrev{local reuse} buffer must be dimensioned to simultaneously
hold all data elements in the {\it localized iteration space}.
Our method is based on the observation that under application control,
%
%
%
different data may be buffered at different loop-nest level, i.e.
there is no one single {\it localized iteration space}.
As a result, our memory performance model results in 
a more accurate buffer size estimation for application-managed
buffers and in better {\it dataflow schedules}.

\section{Memory Performance Model}
\label{model}

Given a limited local reuse buffer capacity, memory performance optimization
consists in finding a {\it dataflow schedule}, i.e.
the computation order, 
\fcrev{
such that \textit{i)} 
the working
set of the computation, fits in the \textit{local reuse buffer};
\textit{ii)} traffic to \fcrev{off-accelerator} memory is minimized, therefore reducing energy
and time dedicated to data movement.
Therefore, using the notation introduced in Section~\ref{background},} building a {\it dataflow schedule} involves specifying the
loop-nest shape via loop tiling and reordering and choosing a
{\it buffering level} for each array reference.

\asrev{
In this section, we develop an analytical model for evaluating the dataflow
schedules that is more accurate than previously published models in
the case of application-managed buffers.
Given a {\it dataflow schedule}, our analytical model computes the local buffer
size and the number of bytes accessed from the \fcrev{off-accelerator} memory required for
this schedule execution.
\fcrev{Note that though we develop our model in a context of the CNN convolution
computation}, \fcrev{in itself it is generic and} applicable to \fcrev{other types of computation organized in loop-nests}.
}

\subsection{Local reuse buffer size}
\label{sec:local_mem_size}

\begin{table}[tb]
\centering
\begin{tabular}{|l|c|c|c|c|c|c|}
\hline
       & \loopname{LFX} & \loopname{LFY} & \loopname{LSX} & \loopname{LSY} & \loopname{LIF} & \loopname{LOF} \\
\hline
$I$    &   1 or R\footref{FN1}   &   R or 1\footref{FN1}   &   1 or R\footref{FN1}   &   R or 1\footref{FN1}   &   1  &   M   \\
$W$    &   1   &   1   &   E   &   E   &   1  &   1   \\
$O$    &   R   &   R   &   1   &   1   &   C  &   1   \\
\hline
\end{tabular}
\vspace{5pt}
\caption{\label{Tab9} \asrev{Reuse distance} of arrays in CNN convolution loop-nest.}
\end{table}
\footnotetext{The reuse of array $I$ is carried by a combination of two
  loops as explained earlier: which pair of loops -- (\loopname{LFX},\loopname{LSX}) or (\loopname{LFY},\loopname{LSY}) -- is carrying the reuse depends on relative ordering of
  the loops.
  \label{FN1}}

\asrev{
Let us first introduce the concept of the \textit{footprint} necessary to
build the analytical model for a loop-nest.
As explained earlier, the \textit{reuse distance} of an array reference with
respect to
a loop corresponds to the number of iterations during which the
corresponding array element is used by the computation. Table \ref{Tab9}
lists the \textit{reuse distances} of CNN convolution array references with
respect to all loops in the loop-nest.
}
The {\it footprint} of an array inside a loop is the portion of array
that is touched while iterating over the loop.
Thus, the {\it footprint} of array $X$ in loop
\loopname{L}  measures the number of distinct elements of $X$ used inside \loopname{L}.
%
%
\fcrev{
Assume $\text{\loopname{L}}_0 \dots \text{\loopname{L}}_{\mathcal{N}-1}$ are the loops in the current dataflow schedule, ordered from the innermost to the outermost one.
If we call $n(\text{\loopname{L}}_i)$ the number of iterations of loop $\text{\loopname{L}}_i$, and
$d_X(\text{\loopname{L}}_i)$ the \asrev{\textit{reuse distance}} of the array reference $X$ with respect
 to loop $\text{\loopname{L}}_i$, the {\it footprint} $F_X(\text{\loopname{L}}_i)$ of array $X$ in loop $\text{\loopname{L}}_i$ can be
computed as follows:
}
\begin{equation}
\label{eq:A}
F_X(\text{\loopname{L}}_i) = F_X(\text{\loopname{L}}_{i-1})\cdot \frac{n(\text{\loopname{L}}_i)}{d_X(\text{\loopname{L}}_i)},
\end{equation}
with $F_X(\text{\loopname{L}}_{-1}) \triangleq 1$.
\fcrev{Intuitively, this means that the footprint of an array $X$ over a given loop $\text{\loopname{L}}_i$ is
the footprint of the same array over 
\asrev{one loop iteration}, multiplied by the
number of times 
\asrev{that elements of}
this array 
\asrev{have} to be replaced in 
\asrev{the local buffer}
during the iterations of $\text{\loopname{L}}_i$.}

%
%
\asrev{
The \textit{footprint} takes into account any reuse of data elements that
exists in a loop. Thus, 
in order for the elements of array $X$ to be reused across iterations of
a loop $\text{\loopname{L}}_i$, the local buffer must be big enough for
holding the full \textit{footprint} of one iteration
of the loop.
Furthermore, if the loop does not carry reuse of the array,
the application-managed buffer can be shared by data elements from multiple loop
iterations.
This means that the actual required size for the application-managed buffer
is computed as follows:
}
\begin{equation}
\label{eq:B}
  B_X(\text{\loopname{L}}_i) = \left\{
  \begin{array}{ll}
    F_X(\text{\loopname{L}}_{i-1})\quad&\text{if $\text{\loopname{L}}_i$ carries $X$} \\
    B_X(\text{\loopname{L}}_{i-1})\quad&\text{if $\text{\loopname{L}}_i$ does not carry $X$} \\
    \end{array}
  \right.
\end{equation}
%
%
\asrev{
Given a reordered and tiled loop-nest, Equation~\ref{eq:B} allows to
recursively compute local buffer requirements for all array references,
starting from the innermost loop in the loop-nest.
Therefore, it allows to evaluate the buffering requirements of a dataflow
schedule, given its loop-nest shape with \textit{buffering levels} annotated
for all data references.
}


\subsection{\fcrev{Off-accelerator} memory traffic}
\label{sec:main_mem_traffic}

\fcrev{Equation~\ref{eq:B} allows to 
\asrev{
evaluate which dataflow schedules are feasible given a certain buffer size,
as well as to compare them based on the minimum local buffer size they
require,
}
but does not give any indication on
its quality in terms of \fcrev{off-accelerator} memory traffic.
}
\fcrev{Let us call $T_X$ the memory traffic computed as number of bytes accessed in \fcrev{off-accelerator} memory for array $X$, and $P_X$ the numerical precision (in bytes) used for its storage.
\asrev{At}
\textit{buffering level} $\text{\loopname{L}}_i$, then the number of 
\asrev{memory} accesses \asrev{to $X$}
is given by the footprint of $X$ 
\asrev{with respect to} $\text{\loopname{L}}_i$ multiplied by the total number of times that loop $\text{\loopname{L}}_i$ is executed:}
\begin{equation}
T_X = P_X \cdot F_X(\text{\loopname{L}}_i) \cdot \prod_{j=i}^{\mathcal{N}-1} n(\text{\loopname{L}}_j)
\label{eq:traffic_x}
\end{equation}
\fcrev{Note that $T_X$ does not depend explicitly on the size of the local buffer, but only on the dataflow schedule, through the footprint $F_X(\text{\loopname{L}}_i)$ and the iterations of the outermost loops.
Whether the schedule fits 
\asrev{with respect to a given} local buffer 
\asrev{capacity} depends only on Equation~\ref{eq:B}.
}

\fcrev{Memory accesses to the $O$ array constitute a special case as they 
\asrev{include} two distinct contributions:
$E \times E \times M$ writes of the final fully computed output feature maps, and
memory accesses corresponding to the accumulation of intermediate partial results (each accumulation composed of two accesses, 1~write + 1~read).
Storage of accumulated partial results typically uses a different (higher) 
\asrev{numerical} precision than that used for the final $O$ array.
}
The total \fcrev{traffic} is therefore the sum of the traffic of the three $I$, $W$, $O$ arrays:
\begin{equation}
T = T_I + T_W + T_{O,\mathrm{acc}} + T_{O,\mathrm{final}}.
\label{eq:traffic_tot}
\end{equation}
\fcrev{
Equations~\ref{eq:traffic_x} and~\ref{eq:traffic_tot} enable a quantitative comparison of dataflow schedules in terms of memory traffic, which is known to be strongly correlated with energy consumption, and of course with system cost~\cite{SzeCYE17,ContiSchBen2018}.
}

\subsection{Dataflow schedule selection procedure}
\label{sec:dataflow_selection}
\asrev{
We conduct the CNN convolution design space exploration in two steps.
In the first step, we compute local buffering requirements for each array
reference at different loop levels across an enumeration of different
loop orders and loop tile sizes.
This step is independent of a particular CNN layer shape because the
local buffering requirements at any loop-nest level depend only on
loop order and tile sizes of different loops.
In the second step, using these pre-enumerated buffer requirements, we
analyze a particular CNN layer, exhaustively searching for a
best combination
of buffering levels for the three CNN arrays under different local buffer capacities.
}

\asrev{
The first step requires the enumeration of 6! = 720 loop-nest permutations.
However, we can reduce this number by not
considering permutations between the two kernel loops (\loopname{LFX} and
\loopname{LFY} in Figure \ref{Fig3}), and between the two image loops 
(\loopname{LSX} and \loopname{LSY}
in Figure \ref{Fig3}). These permutations can be omitted without
affecting our conclusions because they result in symmetric dataflow
schedules.
In order to reduce the enumeration size further, tile sizes are enumerated
selectively. We want to quantify how different loop orderings and tile
sizes affect the number of memory accesses. For this it is not necessary
to enumerate all possible tile sizes; instead we examine a sequence of monotonically increasing, power of two,
tile sizes as well as tile sizes that correspond to common CNN layer
configurations.
The first step results in buffering requirements for each CNN array
reference at each loop level for different
loop-nest shapes.
}

\asrev{
With above search space reduction, the first exploration step yields 180
possible convolution loop-nest
permutations with multiple tiling shapes each.
In the second step, we search within this pre-enumerated loop-nest
permutation
space for \textit{dataflow schedules} that fit the local reuse buffer
capacities between 1KB and 512KB, while minimizing required
\fcrev{off-accelerator} memory access bandwidth.
This search results in one best \textit{dataflow schedule}, i.e. loop-nest
order, tiling sizes, and \textit{buffering levels} for CNN memory references,
for each evaluated convolution layer and for each buffer capacity.
}

\subsection{Comparison vs existing models}

\begin{figure*}[t]
\centering
\includegraphics[width=\linewidth]{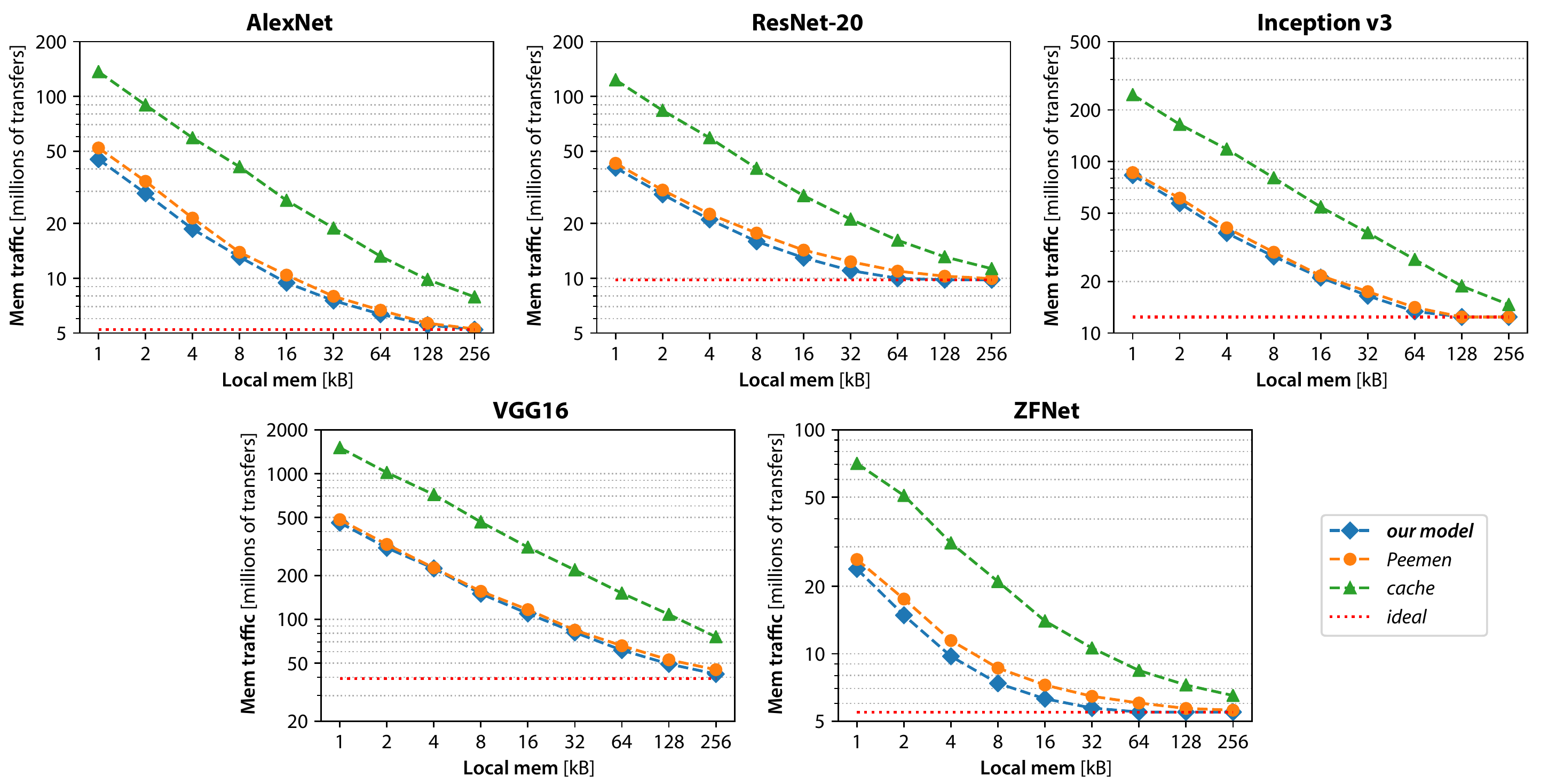}
\caption{\label{fig:all_traffic} \fcrev{Memory traffic comparison of the best dataflow schedules identified by our model, the model proposed in Peemen~\textit{et~al}. and the cache model on a set
  of representative CNNs, while sweeping the local memory constraint from 1 to 256~KB. Both axes are in logarithmic scale. Results aggregate traffic contributions from all convolutional layers; the dashed red line indicates the ideal memory traffic when all data reuse is fully exploited.}}
\end{figure*}

\begin{figure}[t]
\centering
\includegraphics[width=0.99\columnwidth]{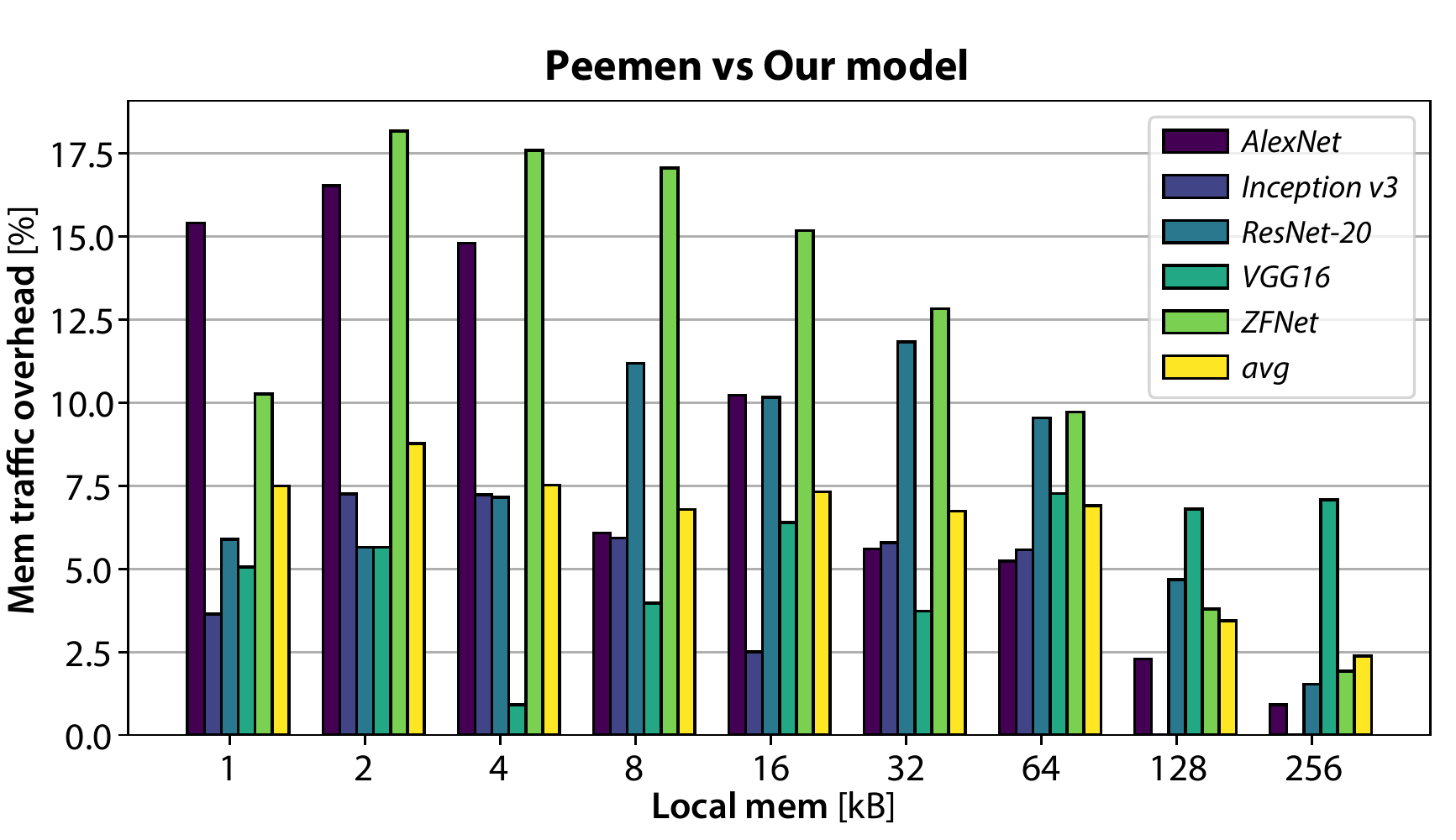}
\caption{\label{fig:peemen_vs_us} \fcrev{Memory traffic overhead with respect to our proposed model when using the model proposed in Peemen~\textit{et~al.} to select the best dataflow schedule, while sweeping the local memory size from 1 to 256 kB.}}
\end{figure}

\fcrev{The main use of the memory performance model is to evaluate the quality of \textit{dataflow schedules} pending a set of local buffering constraints, to determine which schedule minimizes memory traffic.
Therefore, to compare our model with the current state-of-the-art, we first investigated whether it can identify better dataflow schedules than those that can be extracted by previously published models~\cite{WolfLam91,PeemenMC2013,ZhangLSGXC15,YangPRBRKRPH16}.
}
\fcrev{
We selected two representative models, \textsc{cache} and \textsc{Peemen}, to compare with our own proposed model.
}
\fcrev{
Similarly to what happens with our model, the selection of the best dataflow schedules involves
\asrev{exhaustive search over a large solution
space, considering all possible loop tile sizes and loop orderings. }
}

\subsubsection{\textsc{cache}} 
M.~Wolf~{\it et~al.}~\cite{WolfLam91} were first to
apply loop-nest reordering and tiling in order to find the {\it localized iteration space}. 
\asrev{
The method conservatively assumes that each data element in the 
{\it localized iteration space} working set needs to be allocated a place in 
the cache, thus overestimating the required memory footprint. It is also assumed 
that the entire {\it localized iteration space} working set needs to be transferred 
between the memory and the cache for each new {\it localized iteration space}
execution because it cannot be guaranteed that data from a previously execution 
is still present in the cache.
}
\asrev{
The original paper \cite{WolfLam91} also proposed a heuristics for trimming the number of
tiling possibilities, guided by the cache behavior in scientific computations.
}

\subsubsection{\textsc{Peemen}}
\asrev{
The memory performance model proposed by Peemen {\it et al.}~\cite{PeemenMC2013} and
independently by Zhang {\it et al.}~\cite{ZhangLSGXC15}, improves on the
original cache model significantly by taking into account that with application
managed buffers some data can be reused between consecutive executions of the
\textit{localized iteration space}, which they call the \textit{innermost tile}.
By changing the order of controlling loops, different quality solutions are obtained.
Yang {\it et al.}~\cite{YangPRBRKRPH16} published a similar model extended for optimal 
CNN loop-nest tiling for multiple levels of memory hierarchy.
As explained in section \ref{related}, the models developed by Zhang {\it et al.} 
and by Yang {\it et al.}
yield dataflow schedules essentially similar to the schedules
generated by the Peemen {\it et al.} method, therefore we implemented the
latter as a representative model for \asrev{the three} approaches\footnotemark.
\footnotetext{
Appendix \ref{app:A} describes Peemen's memory performance model
that we derived for the loop-nest from the Figure \ref{Fig3}.
For a detailed explanation of these formulas the reader is referred to \cite{PeemenMC2013}.
}

\asrev{
In all these methods,
}
the ordering of the loops inside the \textit{innermost tile} is not
important because the ability to reuse data depends solely on the total number of data loaded
to the local reuse buffer between reuses.
In Peemen's and Zhang's models, only the innermost controlling loop affects the data reuse across 
consecutive iterations of the \textit{innermost tile}. Therefore, fewer loop-nest
permutations need to be explored considerably reducing the solution search space.
}

We have compared the \textit{dataflow schedules} generated by our model with
\textit{dataflow schedules} computed by previously published models \cite{WolfLam91},
\cite{PeemenMC2013}, \cite{ZhangLSGXC15}, and \cite{YangPRBRKRPH16}
\fcrev{over the convolutional layers from five state-of-the-art CNN topologies:}
\asrev{
AlexNet, ZFNet, VGG16, Inception v3, and ResNet-20\footnote{
Configurations of the CNN layers that we've used in our evaluations are
listed in the Appendix \ref{app:B}.}.
Although the following comparison is based on generated dataflow schedules,
we show in
Section \ref{architecture} that our method is exact with respect to the
real execution. Therefore, this estimation corresponds to the actual
amount of memory traffic generated by these CNN layers.
}

Figure \ref{fig:all_traffic} plots the
\asrev{total number of data transfers} to and from
\fcrev{off-accelerator} memory, \fcrev{using the best dataflow schedules}
estimated by \fcrev{the three models, while sweeping the size of the local reuse buffer from 1 to 256 KB.
We aggregate the memory traffic from the best dataflow schedules identified for each layer, and we also show the ideal result given by the ``essential'' memory traffic that is present when all data reuse is exploited.
}
From the plot, it is clear that the cache model largely overestimates
the \asrev{memory traffic} 
requirements compared to the two application managed buffer models,
and always results in sub-optimal dataflow schedules for
\asrev{any buffer size and for}
all CNN convolution layers that we
tested, \fcrev{by a factor of up to 3.5$\times$}.
%

\fcrev{
The advantage of our model when compared with Peemen's method is more subtle, as both exploit the characteristics of application-managed buffers to yield a better schedule.
To analyze the difference between the two models, Figure \ref{fig:peemen_vs_us} plots the relative overhead in memory traffic of Peemen's model with respect to our model.
}
\asrev{
Figure \ref{fig:peemen_vs_us} plots the bandwidth requirements estimated
from Peemen's model as a percentage overhead compared to our model, given
local reuse buffer
capacities between 1KB and 256KB, for the same CNN convolution layers.
Our model finds dataflow schedules with between 2.5\% and 17.5\% lower
memory traffic, with over 10\% memory traffic reduction for several CNNs, especially when targeting
smaller local reuse buffer sizes.
Even for a relatively large local buffer size of 128KB \asrev{and 256KB}, our
method results
in dataflow schedules with more than 5\% memory traffic reduction over the
set of convolution layers for several CNN networks.
%
}

\asrev{
For most of evaluated CNN convolution layers, our method results in some
reduction of memory traffic due to its ability to exploit data reuse across
all levels of the CNN convolution loop-nest.
With rare exceptions, Peemen's method is able to find similar schedules
only when a full volume of the input or the output feature maps can be
stored in the local reuse buffer.
We have noticed the following points that contribute to these results:
\begin{itemize}
\item Our method's footprint calculation better takes into account the data
  reuse because it considers independent buffering for the I, W, O arrays.
  As a result, our buffer requirements are systematically lower for a given
  loop-nest shape, and allow room for bigger tiles to be placed in local
  reuse buffers. 
\item Due to independent buffering of the 3 arrays, our method always places
  memory transfers optimally with respect to the total memory traffic.
\item In Peemen's method, unless \loopname{LTIF} is the innermost controlling
  loop, the memory traffic for the $O$ array is multiplied by 2 to account
  for 1 read and 1 write of partial accumulations. This happens even when
  the \loopname{LIF} loop is not tiled.
\end{itemize}
}

\asrev{
Moreover, for a given CNN convolution layer, the memory traffic
overhead from Peemen's method does not necessarily decrease when
the local reuse buffer capacity is increased.
Increasing the local buffer capacity allows, in the first place, to generate
larger tiles. However, at some buffer capacity points, our method is able
to find a completely different loop ordering and buffering levels for the
data references, such that more important memory traffic reduction can be
achieved than by simply increasing the tile sizes.
}

It is worth noticing that
\asrev{
Yang's algorithm can be modified to achieve the same quality dataflow
schedules as the ones using our approach.
}
It can be verified that applying a 4-level blocking at each memory
hierarchy level - \asrev{essentially enumerating the permutations of the
4 loops, \loopname{LSX}, \loopname{LSY}, \loopname{LIF}, and \loopname{LOF}
from the Figure \ref{Fig3}}, would lead
to schedules equivalent to our method.
\asrev{
However, such 4-level blocking is quite computationally expensive: 
Yang {\it et al.} reported that a 4-level blocking of a single CNN layer
takes 24 hours on a Xeon E5645 processor.
}
\asrev{
Our method is significantly faster: the first step described in Section~\ref{sec:dataflow_selection} is performed only once for many
different CNN layers, whereas in Yang's method, a solution search needs to be performed
for each convolution layer individually.
Even with densely sampled tiling space, for example exploring all tile sizes multiple of 2,
our first step takes $\sim$2 minutes per layer on a
simple Intel i7 processor running at 3,4 GHz,
}

\section{Case Study: ASMP Cluster Accelerator}
\label{architecture}

\fcrev{As a case study for our proposed model, in this Section}
we illustrate 
\fcrev{how it can be used }
for a practical implementation of 
\asrev{a low-cost} CNN accelerator.
\asrev{
\fcrev{Specifically, we } (1) show how our model can be used to derive a \textit{dataflow schedule}
for a specialized hardware block, dedicated for processing CNN convolution
layers, (2) show that
this \textit{dataflow schedule} is implementable, and (3) evaluate
its efficiency in terms of memory bandwidth utilization and accuracy.
}

\begin{figure}[!th]
\centering
\includegraphics[width=0.97\linewidth]{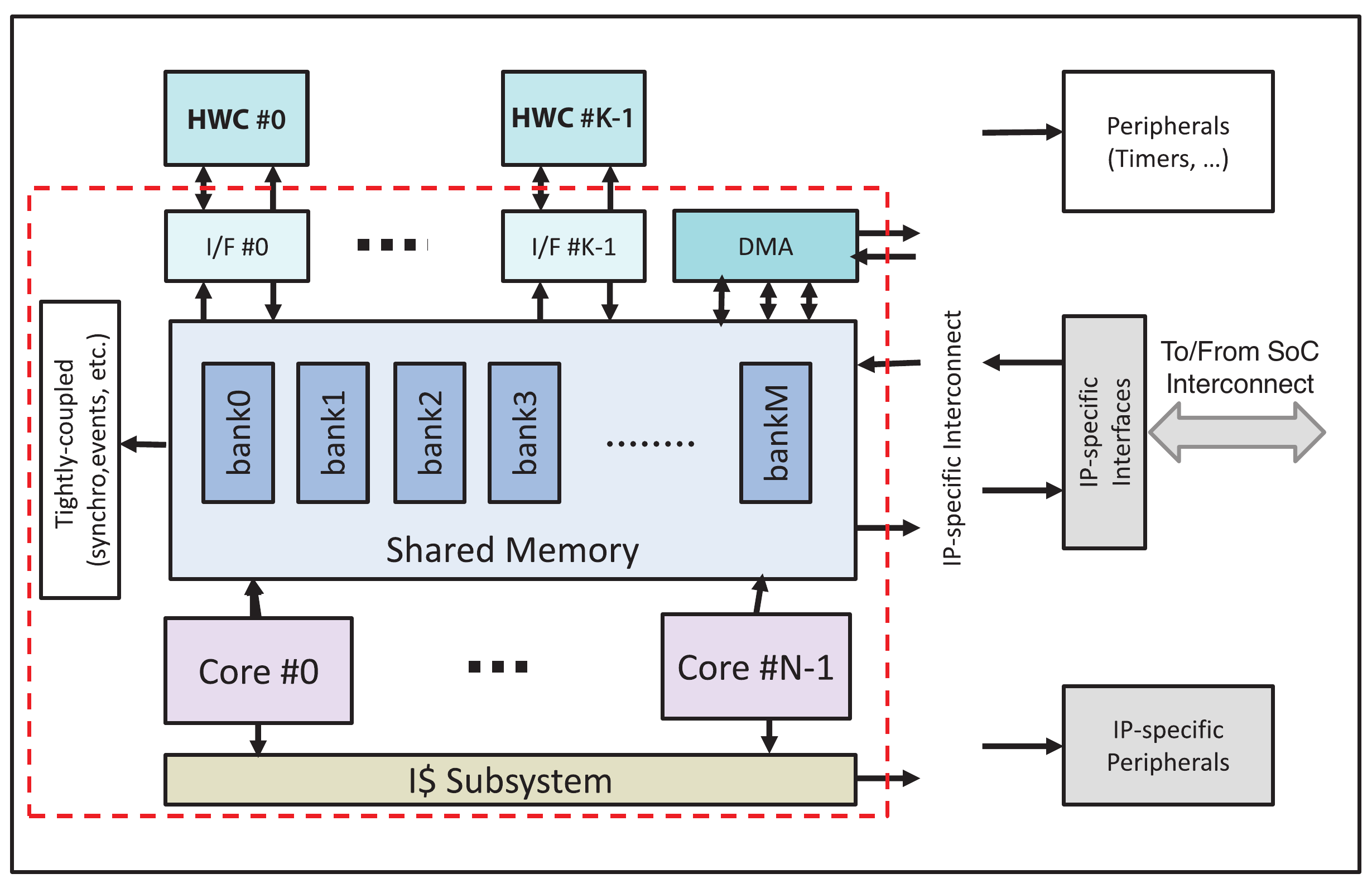}
\caption{\label{asmp} \fcrev{ASMP Accelerator Platform tightly-coupled shared-memory cluster}.}
\end{figure}

\subsection{Target architecture}

\fcrev{As a target architectural template, we chose to integrate special-purpose \textit{convolution hardware blocks} (HWCs) inside STMicroelectronics' ASMP tightly-coupled shared-memory cluster~\cite{BeniniAl2012p2012}.}
Fig. \ref{asmp} shows the block diagram of this architectural template.
An \fcrev{ASMP} cluster is 
composed by a number of programmable cores, HWCs, and a DMA, all connected together to a shared
Tightly-Coupled Data Memory (TCDM) via a single-cycle  {\it logarithmic}
interconnect~\cite{RahimiAl2011}.
\asrev{
The shared memory enables efficient exchange of data between the
convolutional units and programmable processors,
achieving high degree of flexibility
by computing non-convolution functions,
such as pooling,
normalization, etc., in software.
Additionally, such shared memory cluster can efficiently
support traditional computer vision algorithms, ORB, HOG, etc. because
many image processing algorithms are essentially based on convolutional
operations.
}

The current ASMP cluster implementation targets mobile image processing
applications and includes up to 16 RISC processor cores and HWC blocks
running at \fcrev{moderate} frequency (500 MHz), and up to 256KB of the TCDM memory.
The interconnect can be designed with 32-bit or 64-bit TCDM access width
with maximum peak bandwidth of 64 GB per second.
The key element of the ASMP cluster is the {\it logarithmic}
interconnect
that allows multiple concurrent accesses to the multi-bank TCDM
memory.
The logarithmic interconnect provides a common infrastructure for the
core-to-core, the core-to-hardware-block, 
or the hardware-block-to-hardware-block communication.

The \asrev{specialized convolution hardware blocks, the} HWCs, are essential for achieving the required
performance while keeping the cost and the power consumption low.
For designing the HWC, we have leveraged on shared memory dedicated hardware
blocks methodology similar to the one described in \cite{ContiMPB16}.
The tight coupling of the HWC to shared memory allows usage
of complex memory access patterns, such as sliding kernels, repeated
re-fetch of data, etc. This flexibility enables efficient implementation of
vast variety of dataflow schedules.
We have chosen to implement a Single Instruction Multiple Data (SIMD) type of
datapath. The SIMD processing ensures a steady datapath utilization
independent on the convolution kernel size. Furthermore, we have opted for a
relatively narrow 16-byte SIMD width such that the HWC efficiency is maintained
even for smaller images.

\subsection{HWC Design Space Exploration}

\asrev{
The HWC is dedicated to processing the CNN
\textit{convolutional layer}, which accounts for most of the computational work and most of the \fcrev{partial results} data bandwidth in existing CNNs~\cite{SzeCYE17}.
\fcrev{A critical point in the design of HWCs, like most accelerators, is what data has to be internalized within a local buffer and what is accessed from outside the accelerator, in this case from the cluster shared TCDM.
This problem is readily mapped to the conceptual view of our model (Figure~\ref{Fig14}), where internalized memories constitute the \textit{application-managed local reuse buffer}, whereas the TCDM is the \textit{\fcrev{off-accelerator} memory}.}
It is therefore straightforward to apply the memory performance model presented in Section~\ref{model} as a tool for
design space exploration, with the objective to find the best
trade-off
between the HWC local storage capacity and the required TCDM bandwidth.
}
 
\fcrev{Local storage and TCDM bandwidth are generally conflicting objectives.}
\asrev{
On the one hand, 
minimizing local storage capacity is important \fcrev{to reduce} area and therefore cost of the HWC IP.
Our shared cluster platform includes several HWC and a TCDM memory for
buffering data on-chip. Making the total HWC internal storage capacity close to the TCDM capacity~\fcrev{would make its usage redundant, as the access energy for local memory would be comparable to that of a TCDM access.}
%
On the other hand, 
the bandwidth to cluster shared memory remains a scarce resource
because multiple actors in the system are accessing it simultaneously; 
without any local storage at all, every data access from HWC
would be done to the TCDM memory. The resulting bandwidth requirement \fcrev{would exceed} the local
interconnect capacity, leading to a drop in performance and to high energy
consumption.
Furthermore, accessing
the cluster shared memory is 
more expensive in terms of
energy consumption than accessing internal HWC storage~\cite{ContiSchBen2018},
\fcrev{and the number of ports used to connect the HWCs to the TCDM has a significant impact on its size and the maximum working frequency of the cluster -- which means that minimizing TCDM bandwidth requirements is also important.}}

\fcrev{In general, the trade-off between local storage and memory bandwidth} depends \fcrev{in on the shape of the specific} CNN layer: convolution
kernel size, number and size of the feature maps, feature map and kernel
numeric precision.
Therefore, \fcrev{for a general-target HWC} we want to build a CNN loop-nest
\textit{dataflow schedule} that, given a local reuse buffer capacity in the
order of a few KBs, minimizes the TCDM memory \fcrev{traffic -- and therefore bandwidth --} across a large number of CNN layers \fcrev{taken from different CNNs}.
\asrev{
We conducted \fcrev{this} design space exploration in two steps as described in Section \ref{model}.
We analyzed \fcrev{71} different representative CNN layers chosen
from the AlexNet,
ZFNet, VGG, Inception v3 and  ResNet topologies,
exhaustively searching for a best dataflow schedule
for each of 180 different loop-nest permutations under different local buffer capacities.
}

\begin{figure}[!t]
\centering
\includegraphics[width=\linewidth]{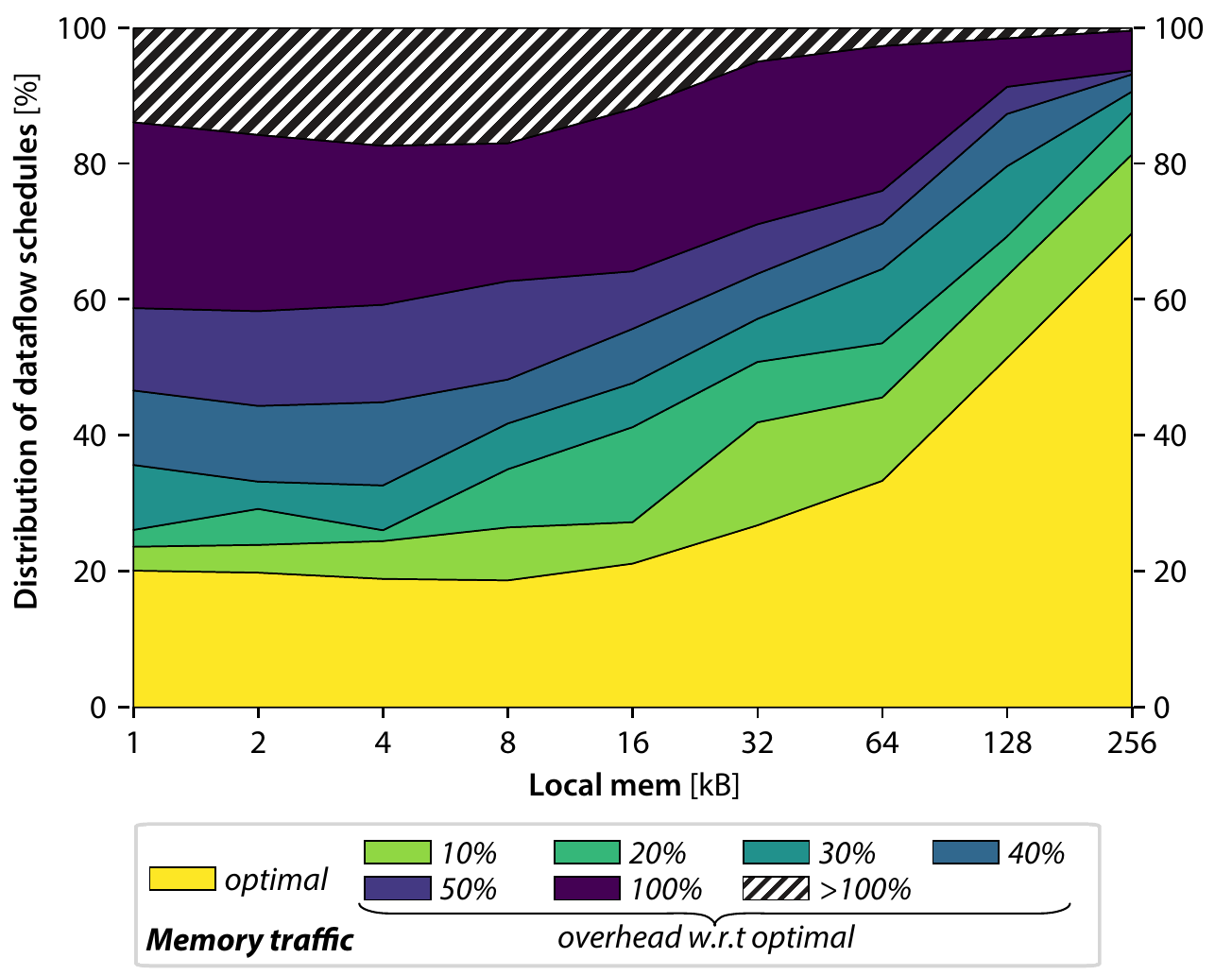}
\caption{\label{Fig6} \fcrev{Distribution of dataflow schedules across different loop-nest permutations, binned depending on the amount of memory traffic they generate.}}
\end{figure}

Figure \ref{Fig6} shows the distribution of the
dataflow schedule quality across the 180 loop-nest
permutations \fcrev{obtained with different local memory constraints, binned according to the amount of memory traffic they generate.}
The Y-axis shows, for each local buffer capacity, the
percentage of loop-nest permutations that result in optimal \fcrev{traffic}, or
\fcrev{add up to} 10\%, 20\%, etc. of \fcrev{overhead to} the optimal \fcrev{traffic}, or exceed 2 times
optimal \fcrev{traffic}, respectively.
For small local buffer capacities, less than 20\% of loop-nest
permutations can achieve optimal bandwidth. On the other hand, with a large
local buffer $\sim$50\% of loop permutations can be tiled in such a
way that optimal bandwidth is achieved.

\lstset{style=kpnapi,numbers=right,stepnumber=1,escapechar=@}

\begin{figure}[!t]
\centering
\begin{minted}[numbersep=5pt, gobble=2, frame=lines, fontsize=\footnotesize, framesep=2mm]{C}
  LTOF: for (mm = 0; mm < M; mm += mss)
    LTIF: for (cc = 0; cc < C; cc += css)
      LTSY: for (yy = 0; yy < E; yy += iss)
        LTSX: for (xx = 0; xx < E; xx += jss)
          // Buffering O
          LIF: for (c=cc;c<min(cc+css,C);c++)
            // Buffering I
            LSY: for (y=yy;y<min(yy+iss,E);y++)
              LFY: for (k = -R/2; k < R/2; k++)
                // Buffering W
                LOF: for (m = mm; m < mss; m++)
                  LSX: for (x=xx;x<min(xx+jss,E);x++)
                    LFX: for (l = -R/2; l < R/2; l++) {
                      p = I[c][y*S+k][x*S+l]
                      w = W[m][c][k][l]
                      O[m][y][x] += p*w
                    }
\end{minted}
\caption{\label{Fig9} HWC dataflow schedule: gives best bandwidth
trade-off with local storage capacity less than 4KB.}
\end{figure}

\fcrev{By analyzing the best dataflow schedules obtained in this experiment, we confirmed the intuition that} those \textit{dataflow schedules} that allow the output
feature maps to be fully accumulated locally by buffering the
partial sums \fcrev{tend to be the best performing ones, especially when the local buffer capacity is small}.
Although in these schedules the
input feature maps and weights are read from memory multiple times, they
still result in fewer total
memory accesses compared to dataflow schedules where the output feature
maps are swapped out to memory before being fully
accumulated through all of the input feature maps. Swapping and
re-fetching the output feature maps to complete the accumulation generates
twice the traffic
compared with the read-only input feature maps and weights.
Furthermore, the partially accumulated output feature maps require higher
precision and therefore are more costly in terms of required bandwidth.
It is interesting to notice that given less than 512KB buffering
capacity, no one permutation resulted in a dataflow schedule with optimal
bandwidth across the
entire set of tested convolution layers.

\asrev{
Among several small footprint schedules, we chose one that, for most
tested CNN layers, leads to
minimal memory bandwidth requirements for local buffer sizes from 1KB
to 4KB.
Our selection was also guided by several hardware implementation
}
criteria, such as the number of required simultaneous local buffer accesses,
access alignment, etc. \fcrev{and the compatibility with a SIMD datapath}.
Figure~\ref{Fig9} shows the dataflow schedule chosen for the HWC
implementation, \fcrev{with the buffering} level
for each array shown as a comment on top of the corresponding loop.
The HWC main loop executes an innermost tile in order  \loopname{LIF} -  \loopname{LSY} -  \loopname{LFY} - \loopname{LOF} -  \loopname{LSX} -  \loopname{LFX}.
\fcrev{The relative order of} loops  \loopname{LSX},  \loopname{LFX} and  \loopname{LFY} ensures that the partial sum accumulation
remains internal to the HWC as much as possible.
In the actual implementation, the tiling factor $jss$ for the  \loopname{LIF} loop is
fixed and equals the SIMD datapath width.
The remaining tile dimensions: the
number of output feature maps, $mss$, the input feature maps, $css$, and
the number of output lines in a tile, $iss$, are computed for each
particular convolution layer \asrev{also} using our performance model.
In practice, over all tested CNN layers, the input feature maps volume was
never
tiled ($css = C$), allowing a complete accumulation of the partial sums
inside the HWC local buffer.

\subsection{HWC evaluation}

\begin{figure}[!th]
\centering
\includegraphics[width=0.8\linewidth]{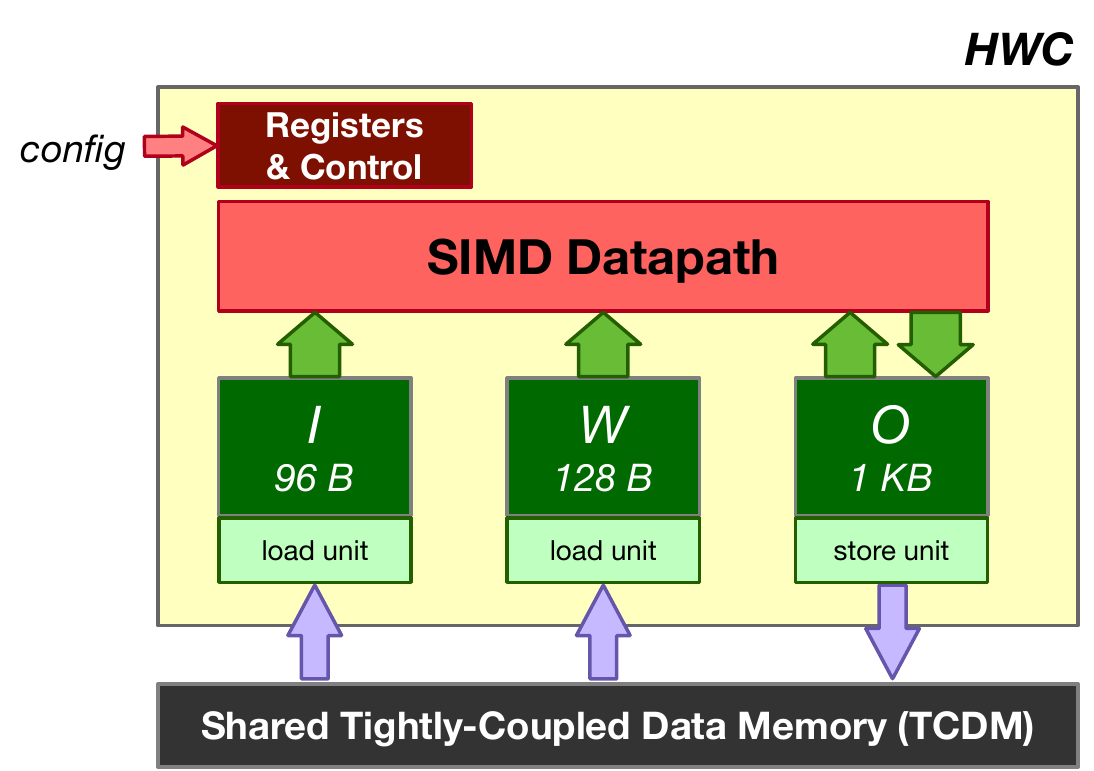}
\caption{\label{HWC} HWC block internal architecture.}
\end{figure}

\fcrev{To explore the hardware implications of the \textit{dataflow schedule} proposed in Figure~\ref{Fig9}, we specified a HWC prototype using it directly in C, using the CatapultC high-level synthesis (HLS) tool to derive a Verilog design.
The design exposes one slave configuration port and 3$\times$32-bit master ports towards TCDM (one for each array reference) to separate control for memory accesses for each array, easing the high-level synthesis process.
}
\fcrev{We specified the HWC datapath so that it is capable of handling} convolution kernel sizes up to
11$\times$11 with stride and unlimited input/output feature map sizes.
\fcrev{It is design as single-instruction multiple data (SIMD) engine and it} is capable of \fcrev{sixteen} 8-bit$\times$8-bit or 8 
16-bit$\times$16-bit fixed-point MAC
operations per clock cycle.
\asrev{
Figure \ref{HWC} shows the block diagram of the HWC and the final breakdown of the local storage capacity for
the three CNN arrays.
\fcrev{The prototype HWC} has slightly over 1KB of internal storage including small input ($I$),
weight ($W$), and accumulated sum ($O$) buffers, with partial sums
accumulated in 32-bit precision.
}


\fcrev{Using the HWC design generated by HLS, we performed synthesis and place-and-route of an ASMP cluster with 4 HWC targeting a 28nm technology node, achieving a maximum frequency of 500 MHz.}
\fcrev{The} computing cluster achieves up to 64 8x8 MAC operations per
cycle for a total of 32 GMAC/s at 500 MHz, with average utilization of 80\%
across the set of convolutional layers \fcrev{used for the design space exploration}.

\asrev{
To understand how accurately our memory performance
model evaluates the memory traffic with respect to an actual implementation,
}
 we have measured the actual number of bytes transferred between the HWC and the TCDM
during execution of various CNN layers
and compared these measurements
with the bandwidth predicted by our memory performance model.
Our memory performance model is almost exact with respect to the measured bandwidth.
\fcrev{The memory performance model does not explicitly account for strided convolutional layers when tile sizes are not integral multiples of the $(iss,jss)$ tile; a small overestimation of the memory traffic can be observed in such cases.}
From our experiments such overestimation is \fcrev{always within} 0.5\% of the total memory traffic.


\begin{figure}[!t]
\centering
\includegraphics[width=\linewidth]{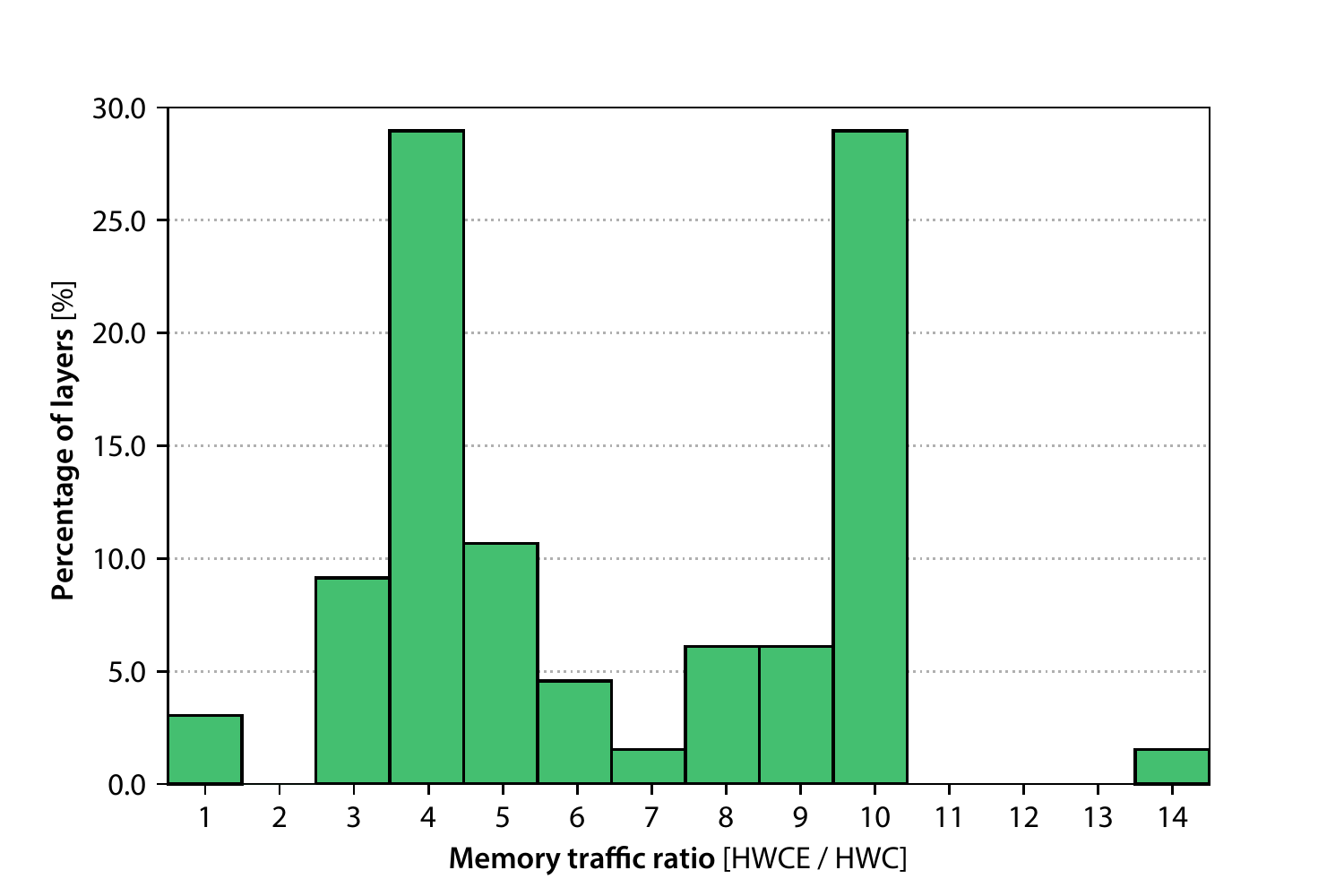}
\caption{\label{Fig7} \fcrev{Histogram of ratios of memory traffic achieved using the HWCE dataflow schedule vs the one proposed for the HWC, under the constraint of a local storage capacity of 1KB.}}
\end{figure}


\asrev{
In order to put the HWC schedule into perspective, we compare the amount
of memory traffic generated by the HWC to the memory traffic generated
by a state-of-the-art tightly-coupled CNN
convolution accelerator, HWCE~\cite{ContiBenini15,ContiAl16}.
Both hardware units target ultra low-cost and low-power applications
and implement very little internal buffer storage.
Both hardware units target tightly-coupled shared memory clusters and
are constrained by the performance of the shared memory logarithmic
interconnect in a similar way.
The HWCE implements a 2D convolution dataflow with linear input buffering
similar to \cite{FarabetPHL09}.
}

\begin{figure}[!t]
\centering
\begin{minted}[numbersep=5pt, gobble=2, frame=lines, fontsize=\footnotesize, framesep=2mm]{C}
  // X-direction, image split in stripes in SW
  LTSX: for (xx = 0; xx < E; xx += jss)
    // output fmaps -- loop tiled in SW
    LTOF: for (mm = 0; mm < M; mm += 1)
      // input fmaps -- loop tiled in SW
      LTIF: for (cc = 0; cc < C; cc += 1)
        // Y-direction -- innermost controlling loop
        LTSY: for (yy = 0; yy < E; yy += iss)
          // Below executed in HWCE
          // Buffering I,W
          LOF: for (m=mm; m<mm+1; m++)
            LIF: for (c=cc; c<cc+1; c++)
              // spatial -- tile loops
              LSY: for (y=yy; y<min(yy+iss,E); y++)
                LSX: for (x=xx; x<min(xx+jss,E); x++)
                  // Buffering O
                  LFY: for (k=0; k<R; k++)
                    LFX: for (l=0; l<R; l++)
                    {
                      p = I[c][y*S+k][x*S+l];
                      w = W[m][c][k][l];
                      O[m][y][x] += p*w;
                    }
\end{minted}
\caption{\label{Fig16} HWCE CNN convolution layer loop-nest}
\end{figure}

\fcrev{As shown in Figure~\ref{Fig16}, the HWCE uses a variant of the canonical tiled loop-nest shown in Figure~\ref{Fig3}, with the outermost tiling loops executed in software -- it executes a 2D convolution
over a single input feature map resulting in one partially computed output
feature map.}
\asrev{
The HWCE takes limited advantage of the data reuse existing
in the convolution loop-nest.
The \textit{convolution reuse} results from the input feature maps
being buffered in an input linear
buffer~\cite{FarabetPHL09}. The \textit{weight reuse} is ensured by buffering
the convolution kernel weights required for processing a single pair of
an input and an output feature map. The output feature maps and the
partially accumulated sums are stored in the TCDM memory, the elements
of the $O$ array are only buffered while applying a single convolution kernel.
}

Figure \ref{Fig7} shows the ratio of the HWCE \fcrev{traffic} for
different CNN layers vs the HWC \fcrev{prototype using 1KB of} internal buffer.
\fcrev{While HWC and HWCE expose the same number of ports towards TCDM (3$\times$32 bits), 
the HWC dataflow schedule results in lower traffic for all
tested convolutional layers.}
For some layers, the HWC dataflow schedule results in up to
14 times reduction in
\fcrev{traffic} compared to the HWCE dataflow.
\asrev{
The HWCE dataflow schedule suffers from important penalty due to
large amounts of
partial accumulation sums that are stored in the TCDM memory.
Writing and reading these partial sums with higher numeric precision
result in noticeable increase in memory traffic.
}
Additionally, with 1KB memory budget, the HWCE linear input buffer would
be too small
for some layers, such as AlexNet-1, ZFNet-1, or ResNet-1 - a bar is absent for
such layers in the Figure.
To handle insufficient linear input buffer size, the actual HWCE
implementation splits the input feature maps
into several smaller stripes that can be handled with the small linear input
buffer -
this results in a further slight increase of redundant shared memory traffic,
due to the overlap between stripes.
\fcrev{Since HWC and HWCE expose the same number of TCDM ports, the ones on the HWC are 
used, on average, significantly less. This translates to a lower amount of energy
spent in transactions with memory and on less contention, making it easier to 
combine HWC operation with other computations in the cluster~\cite{ContiMPB16} without
significant performance hits.
}

\section{Conclusion and Future Work}

We have presented an analytic memory performance model suitable for
memory hierarchies that use application managed buffers.
We have shown that our model results in more accurate memory footprint
estimation than previously published models and is accurate with respect
to a real implementation.
We have used this model for designing a CNN convolution hardware block
in the context of the ASMP shared memory cluster.
Our further work includes applying our model to automatic generation
of dataflow schedules from standard CNN descriptions in
Caffe, TensorFlow or similar tools.







\bibliographystyle{IEEEtran}


\begin{appendices}

\section{Peemen Equations for the CNN Convolution Loop-Nest}
\label{app:A}

\asrev{
The buffering requirements for the 3 array references in the CNN convolution
loop-nest are computed as shown in Equation \ref{eq:C}:
}
\asrev{
\begin{align}
\label{eq:C}
& B(I) = css \times ((iss-1)*S+R) \times ((jss-1)*S+R) \notag \\
& B(W) = mss \times css \times R \times R \\
& B(O) = mss \times iss \times jss \notag
\end{align}
}
\asrev{
The terms $((iss-1)*S+R)$ and $((jss-1)*S+R)$ compute the dimensions, in
pixels, of the input feature map tile, given the tile size for the output
feature map, $iss \times jss$, the convolution kernel size, $R \times R$,
and the convolution stride~$S$.
}

\asrev{
The total memory traffic is computed by multiplying the buffering
requirements by the total number of tiles,
with one improvement.
Peemen noticed that using application-managed buffers, data can also be
reused between consecutive innermost tile executions.
This significantly improves the memory traffic estimation accuracy for
the computations involving prologue, steady state, and the epilogue.
Equation \ref{eq:D} shows the general form of the memory traffic
computation:
}
\asrev{
\begin{equation}
\label{eq:D}
T = \ceil[\bigg]{\frac{M}{mss}} \ceil[\bigg]{\frac{C}{css}} \ceil[\bigg]{\frac{E}{iss}} \ceil[\bigg]{\frac{E}{jss}} (B(I) + B(W) + 2*B(O))
\end{equation}
}
\asrev{
The 4 cases that need to be considered for the CNN loop-nest, corresponding
each to one of the controlling loops,
\loopname{LTSX},
\loopname{LTSY}, \loopname{LTIF}, and \loopname{LTOF}, being the innermost
controlling loop, are shown in Equations \ref{eq:E}, \ref{eq:F}, \ref{eq:G},
and \ref{eq:H} below.
}

\begin{enumerate}
\item Innermost \loopname{LTOF}:
\asrev{
\begin{align}
\label{eq:E}
T^1 = \ceil[\bigg]{\frac{C}{css}} \ceil[\bigg]{\frac{E}{iss}} \ceil[\bigg]{\frac{E}{jss}} (B^1(I) + B^1(W) + B^1(O))
\end{align}
with
\begin{align*}
& B^1(I) = css \cdot ((iss-1)*S+R) \cdot ((jss-1)*S+R) \\
& B^1(W) = M \cdot css \cdot R \cdot R \\
& B^1(O) = 2 \cdot M \cdot iss \cdot jss
\end{align*}
}
\item Innermost \loopname{LTIF}:
\asrev{
\begin{align}
\label{eq:F}
T^2 = \ceil[\bigg]{\frac{M}{mss}} \ceil[\bigg]{\frac{E}{iss}} \ceil[\bigg]{\frac{E}{jss}} (B^2(I) + B^2(W) + B^2(O))
\end{align}
with
\begin{align*}
& B^2(I) = C \cdot ((iss-1)*S+R) \cdot ((jss-1)*S+R) \\
& B^2(W) = mss \cdot C \cdot R \cdot R \\
& B^2(O) = mss \cdot iss \cdot jss
\end{align*}
}
\item Innermost \loopname{LTSY}:
\asrev{
\begin{align}
\label{eq:G}
T^3 = \ceil[\bigg]{\frac{M}{mss}} \ceil[\bigg]{\frac{C}{css}} \ceil[\bigg]{\frac{E}{iss}} (B^3(I) + B^3(W) + B^3(O))
\end{align}
with
\begin{align*}
& B^3(I) = css \cdot H \cdot ((jss-1)*S+R) \\
& B^3(W) = mss \cdot css \cdot R \cdot R \\
& B^3(O) = 2 \cdot mss \cdot E \cdot jss
\end{align*}
}
\item Innermost \loopname{LTSX}:
\asrev{
\begin{align}
\label{eq:H}
T^4 = \ceil[\bigg]{\frac{M}{mss}} \ceil[\bigg]{\frac{C}{css}} \ceil[\bigg]{\frac{E}{jss}} (B^4(I) + B^4(W) + B^4(O))
\end{align}
with
\begin{align*}
& B^4(I) = css \cdot ((iss-1)*S+R) \cdot H \\
& B^4(W) = mss \cdot css \cdot R \cdot R \\
& B^4(O) = 2 \cdot mss \cdot iss \cdot E
\end{align*}
}
\end{enumerate}
\asrev{
In Equations \ref{eq:E} - \ref{eq:H}, in order to account for the data reuse
across consecutive innermost tile executions, the memory traffic computation is done
as if the innermost controlling loop were not tiled.
For example, with loop \loopname{LTOF} being the innermost controlling loop, 
the tile size of this loop is set to the total loop count, i.e. $mss = M$,
inside the general equation \ref{eq:D}.
}

\section{Used CNN Layers Configuration}
\label{app:B}

\begin{table}[!h]
\centering
\begin{tabular}{|l|c|c|c|c|c|}
\hline
      Layer  &  $H \times H$     & $E \times E$    & $C$ & $M$ & Conv. \\
\hline
AlexNet 1    &  $224\times 224$ & $55 \times 55$  & 3   & 96  & $11 \times 11, 4$ \\
\hline
AlexNet 2    &  $55\times 55$ & $27 \times 27$  & 96   & 256 & $5 \times 5, 2$ \\
\hline
AlexNet 3    &  $27\times 27$ & $13 \times 13$  & 256  & 384 & $3 \times 3, 2$ \\
\hline
AlexNet 4    &  $13\times 13$ & $13 \times 13$  & 384  & 384 & $3 \times 3, 1$ \\
\hline
AlexNet 5    &  $13\times 13$ & $13 \times 13$  & 384  & 256 & $3 \times 3, 1$ \\
\hline
ZFNet 1    &  $224\times 224$ & $112 \times 112$  & 3 & 96 & $7 \times 7, 2$ \\
\hline
ZFNet 3    &  $13\times 13$ & $13 \times 13$  & 256 & 384 & $3 \times 3, 1$ \\
\hline
ZFNet 4    &  $13\times 13$ & $13 \times 13$  & 384 & 384 & $3 \times 3, 1$ \\
\hline
ZFNet 5    &  $13\times 13$ & $13 \times 13$  & 384 & 256 & $3 \times 3, 1$ \\
\hline
ZFNet 6    &  $6\times 6$ & $6 \times 6$  & 256 & 256 & $3 \times 3, 1$ \\
\hline
VGG 1    &  $224\times 224$ & $224 \times 224$  & 3 & 64 & $3 \times 3, 1$ \\
\hline
VGG 2    &  $224\times 224$ & $224 \times 224$  & 64 & 64 & $3 \times 3, 1$ \\
\hline
VGG 3    &  $112\times 112$ & $112 \times 112$  & 64 & 128 & $3 \times 3, 1$ \\
\hline
VGG 4    &  $112\times 112$ & $112 \times 112$  & 128 & 128 & $3 \times 3, 1$ \\
\hline
VGG 5    &  $56\times 56$ & $56 \times 56$  & 128 & 256 & $3 \times 3, 1$ \\
\hline
VGG 6    &  $56\times 56$ & $56 \times 56$  & 256 & 256 & $3 \times 3, 1$ \\
\hline
VGG 8    &  $28\times 28$ & $28 \times 28$  & 512 & 256 & $3 \times 3, 1$ \\
\hline
VGG 9    &  $28\times 28$ & $28 \times 28$  & 512 & 512 & $3 \times 3, 1$ \\
\hline
VGG 11   &  $14\times 14$ & $14 \times 14$  & 512 & 512 & $3 \times 3, 1$ \\
\hline
\end{tabular}
\vspace{5pt}
\caption{\label{Tab6} Configuration of CNN layers used in our evaluation.}
\end{table}


%
%

\begin{table}[!h]
\centering
\begin{tabular}{|l|c|c|c|c|c|}
\hline
  \#  &  $H \times H$     & $E \times E$    & $C$ & $M$ & Conv. \\
\hline
0   &  $35\times 35$ & $35 \times 35$  & 192 & 64 &
\(
\begin{array}{c}
  1 \times 1, 1 \\
\end{array}
\)
\\
\hline
&             &           &
\(
\begin{array}{c}
  192 \\
  48 \\
\end{array}
\)
&
\(
\begin{array}{c}
  48 \\
  64 \\
\end{array}
\)
&
\(
\begin{array}{c}
    1 \times 1, 1 \\
    5 \times 5, 1 \\
\end{array}
\) 
\\
\hline
&              &           &
\(
\begin{array}{c}
  192 \\
  64 \\
  96 \\
\end{array}
\)
&
\(
\begin{array}{c}
  64 \\
  96 \\
  96 \\
\end{array}
\)
&
\(
\begin{array}{c}
  1 \times 1, 1 \\
  3 \times 3, 1 \\
  3 \times 3, 1
\end{array}
\) 
\\
\hline
&   &  & 192 & 32 &
\(
\begin{array}{c}
  1 \times 1, 1 \\
\end{array}
\) 
\\
\hline

1   &  $35\times 35$ & $35 \times 35$  & 256 & 64 &
\(
\begin{array}{c}
  1 \times 1, 1 \\
\end{array}
\)
\\
\hline
&             &           &
\(
\begin{array}{c}
  256 \\
  48 \\
\end{array}
\)
&
\(
\begin{array}{c}
  48 \\
  64 \\
\end{array}
\)
&
\(
\begin{array}{c}
  1 \times 1, 1 \\
  5 \times 5, 1 \\
\end{array}
\) 
\\
\hline
&              &           &
\(
\begin{array}{c}
  256 \\
  64 \\
  96 \\
\end{array}
\)
&
\(
\begin{array}{c}
  64 \\
  96 \\
  96 \\
\end{array}
\)
&
\(
\begin{array}{c}
  1 \times 1, 1 \\
  3 \times 3, 1 \\
  3 \times 3, 1
\end{array}
\) 
\\
\hline
&   &  & 256 & 64 &
\(
\begin{array}{c}
  1 \times 1, 1 \\
\end{array}
\) 
\\
\hline

2   &  $35\times 35$ & $35 \times 35$  & 288 & 64 &
\(
\begin{array}{c}
  1 \times 1, 1 \\
\end{array}
\)
\\
\hline
&             &           &
\(
\begin{array}{c}
  288 \\
  48 \\
\end{array}
\)
&
\(
\begin{array}{c}
  48 \\
  64 \\
\end{array}
\)
&
\(
\begin{array}{c}
  1 \times 1, 1 \\
  5 \times 5, 1 \\
\end{array}
\) 
\\
\hline
&              &           &
\(
\begin{array}{c}
  288 \\
  64 \\
  96 \\
\end{array}
\)
&
\(
\begin{array}{c}
  64 \\
  96 \\
  96 \\
\end{array}
\)
&
\(
\begin{array}{c}
  1 \times 1, 1 \\
  3 \times 3, 1 \\
  3 \times 3, 1
\end{array}
\) 
\\
\hline
&   &  & 288 & 64 &
\(
\begin{array}{c}
  1 \times 1, 1 \\
\end{array}
\) 
\\
\hline

3   &  $35\times 35$ & $17 \times 17$  & 288 & 384 &
\(
\begin{array}{c}
  3 \times 3, 2 \\
\end{array}
\)
\\
\hline
&
\(
\begin{array}{c}
  35 \times 35 \\
  35 \times 35\\
  35 \times 35\\
\end{array}
\)
&
\(
\begin{array}{c}
  35 \times 35\\
  35 \times 35\\
  17 \times 17\\
\end{array}
\)
&
\(
\begin{array}{c}
  288 \\
  64 \\
  96 \\
\end{array}
\)
&
\(
\begin{array}{c}
  64 \\
  96 \\
  96 \\
\end{array}
\)
&
\(
\begin{array}{c}
  1 \times 1, 1 \\
  3 \times 3, 1 \\
  3 \times 3, 2 \\
\end{array}
\) 
\\
\hline
& $17 \times 17$  & $17 \times 17$ & 288 & 64 &
\(
\begin{array}{c}
  1 \times 1, 1 \\
\end{array}
\) 
\\
\hline

4   &  $17\times 17$ & $17 \times 17$  & 798 & 192 &
\(
\begin{array}{c}
  1 \times 1, 1 \\
\end{array}
\)
\\
\hline
&             &           &
\(
\begin{array}{c}
  768 \\
  128 \\
  128 \\
\end{array}
\)
&
\(
\begin{array}{c}
  128 \\
  128 \\
  192 \\
\end{array}
\)
&
\(
\begin{array}{c}
  1 \times 1, 1 \\
  1 \times 7, 1 \\
  7 \times 1, 1 \\
\end{array}
\) 
\\
\hline
&              &           &
\(
\begin{array}{c}
  768 \\
  128 \\
  128 \\
  128 \\
  128 \\
\end{array}
\)
&
\(
\begin{array}{c}
  128 \\
  128 \\
  128 \\
  128 \\
  192 \\
\end{array}
\)
&
\(
\begin{array}{c}
  1 \times 1, 1 \\
  7 \times 1, 1 \\
  1 \times 7, 1 \\
  7 \times 1, 1 \\
  1 \times 7, 1 \\
\end{array}
\) 
\\
\hline
&   &  & 768 & 192 &
\(
\begin{array}{c}
  1 \times 1, 1 \\
\end{array}
\) 
\\
\hline

\end{tabular}
\vspace{5pt}
\caption{\label{Tab7} Configuration of Inception v3 used in our evaluation.}
\end{table}


%
%

\begin{table}[!h]
\centering
\begin{tabular}{|l|c|c|c|c|c|}
\hline
  \#  &  $H \times H$     & $E \times E$    & $C$ & $M$ & Conv. \\
\hline

1   & $224 \times 224$ & $112 \times 112$ & 3 & 64 & $7 \times 7, 2$ \\
\hline

2   & $56 \times 56$  & $56 \times 56$ &
\( \begin{array}{c}
  64 \\
  64 \\
  64 \\
\end{array}
\)
&
\(
\begin{array}{c}
  64 \\
  64 \\
  256 \\
\end{array}
\)
&
\(
\begin{array}{c}
    1 \times 1, 1 \\
    3 \times 3, 1 \\
    1 \times 1, 1 \\
\end{array}
\) 
\\
\hline

3 & $28 \times 28$ & $28 \times 28$ &
\(
\begin{array}{c}
  256 \\
  128 \\
  128 \\
\end{array}
\)
&
\(
\begin{array}{c}
  128 \\
  128 \\
  512 \\
\end{array}
\)
&
\(
\begin{array}{c}
    1 \times 1, 1 \\
    3 \times 3, 1 \\
    1 \times 1, 1 \\
  \end{array}
\) 
\\
\hline

4 & $14 \times 14$ & $14 \times 14$ &
\(
\begin{array}{c}
  512 \\
  256 \\
  256 \\
\end{array}
\)
&
\(
\begin{array}{c}
  256 \\
  256 \\
  1024 \\
\end{array}
\)
&
\(
\begin{array}{c}
    1 \times 1, 1 \\
    3 \times 3, 1 \\
    1 \times 1, 1 \\
  \end{array}
\) 
\\
\hline

5 & $7 \times 7$ & $7 \times 7$ &
\(
\begin{array}{c}
  1024 \\
  512 \\
  512 \\
\end{array}
\)
&
\(
\begin{array}{c}
  512 \\
  512 \\
  2048 \\
\end{array}
\)
&
\(
\begin{array}{c}
    1 \times 1, 1 \\
    3 \times 3, 1 \\
    1 \times 1, 1 \\
  \end{array}
\) 
\\
\hline

\end{tabular}
\vspace{5pt}
\caption{\label{Tab8} Configuration of ResNet used in our evaluation.}
\end{table}

\end{appendices}

\end{document}